\documentclass[10pt,twocolumn,letterpaper]{article}

\newif\ifarxiv          \arxivtrue   
\ifarxiv
  \usepackage[pagenumbers]{wacv}        
\else
  \usepackage[review,applications]{wacv}  
  \def\wacvPaperID{404} 
  \def\confName{WACV}
  \def\confYear{2026}
\fi
\usepackage{pifont}

\usepackage{graphicx}
\usepackage{booktabs}
\usepackage{multirow}
\usepackage{makecell}
\usepackage{float}
\usepackage[ruled,vlined,linesnumbered]{algorithm2e}
\usepackage[accsupp]{axessibility}  
\usepackage{booktabs}
\usepackage{multirow}
\usepackage{tabularx}
\usepackage{subcaption}
\newif\ifshowcomments          \showcommentstrue    
\newcommand{\name} {ReaDiT}
\ifshowcomments
  \newcommand{\draftcomment}[3]{{\color{#2}\bf [#1: #3]}}
\else
  \newcommand{\draftcomment}[3]{}      %
\fi

\newcommand{\cmark}{\textcolor{green!60!black}{\ding{51}}}
\newcommand{\xmark}{\textcolor{red}{\ding{55}}}

\definecolor{wacvblue}{rgb}{0.21,0.49,0.74}
\usepackage[pagebackref,breaklinks,colorlinks,allcolors=wacvblue]{hyperref}

\newcommand{\nocontentsline}[3]{}
\let\origcontentsline\addcontentsline
\newcommand\stoptoc{\let\addcontentsline\nocontentsline}
\newcommand\resumetoc{\let\addcontentsline\origcontentsline}
\renewcommand{\thefootnote}{\arabic{footnote}}

\title{ReaDiT Guidance: Control for Image and Video Generation using Diffusion Transformer Features}

\author{
Jay Mahajan$^*$\hspace{0.8em}
Chang Liu$^*$\hspace{0.8em}
Rauf Makharov\hspace{0.8em}
Viraj Shah$^{\S}$\hspace{0.8em}
Alexander Schwing\textsuperscript{\dag} \hspace{0.8em}
Svetlana Lazebnik\textsuperscript{\dag} \\
University of Illinois Urbana-Champaign 
}

\begin{document}
\stoptoc
\maketitle

\ifarxiv
  \begingroup
  \renewcommand{\thefootnote}{}
  \footnotetext{\textsuperscript{*} Equal contribution. Main correspondence to Jay Mahajan}
  \footnotetext{\textsuperscript{\S} Currently at Google.}
  \footnotetext{\textsuperscript{\dag} Equal advising}
  \endgroup
\fi

\begin{abstract}
  We present DiT Readout (ReaDiT) Guidance, a lightweight framework for controlling generation with Diffusion Transformer (DiT) models via their internal feature representations. \name{} Guidance uses features from a single DiT block 
  to steer the generative process according to spatial targets -- like depth, pose, or edge maps -- provided at test time. Furthermore, since modern text-to-video models are largely built on DiT backbones, ReaDiT Guidance naturally extends to video generation, enabling camera and motion control. Experimental results demonstrate that our approach 
  achieves competitive or improved results compared to existing feature-based and off-the-shelf adapter-based approaches while requiring fewer parameters. Project page: \url{https://readit-official.github.io/}
\end{abstract}
\section{Introduction}
\label{sec:intro}

Diffusion models have been tremendously successful in generating high-quality images and videos from a given text prompt. 
However, it
is difficult to only use the text prompt to guide generation towards nuanced spatial targets, such as a desired depth-map, pose-skeleton-map, or edge-map. While more recent architectures 
have increasingly supported additional image-based conditioning, the desire and necessity for fine-grained, localized control of the output via other data modalities, such as encouraging the depth, pose, or edge maps of generated subjects, remains of high interest.

Existing plug-and-play solutions for fine-grained, localized control, such as ControlNet \cite{zhang2023adding} and similar adapter networks \cite{mou2023t2i,lin2025ctrl,xuctrlora,duan2025unic}, are highly effective for image generation. Yet, methods like ControlNet require large amounts of paired training data and introduce substantial additional parameters. More recent lightweight alternatives \cite{ominicontrol, zhang2025easycontrol} significantly reduce this overhead, but are designed for specific backbones and image generation. When extending ControlNet and other adapter networks to video generation, these limitations become a significant bottleneck. Scaling up these methods is computationally demanding and data-intensive. Existing lightweight methods that bypass heavy adapters are task specific. For instance, a technique designed for motion control cannot be seamlessly applied to depth conditioning, and vice versa.

To develop a unified method that does not rely on expensive adapters, one can turn to the internal feature representations of the diffusion models themselves. Prior work, such as ``Readout Guidance'' (RG) \cite{luo2024readoutguidancelearningcontrol} , successfully proposed the use of lightweight predictor networks and an inference-time guidance mechanism to achieve controllable generation. However, this foundational method was designed specifically for U-Net-based \cite{podell2023sdxlimprovinglatentdiffusion} diffusion architectures, utilizing features extracted across internal U-Net layers. With the recent paradigm shift towards Diffusion Transformers (DiT) \cite{Peebles2023DiT}, it remains unclear whether such an approach is viable, as the DiT feature space differs fundamentally. 

In this work, we  propose \name{}, an efficient architecture for spatially controlling DiT models via its features. It is based on the observation that, in contrast to U-Net-based models, the internal features from a single transformer block suffice to predict and guide the spatial attributes of a generated image. Based on this observation, we develop a lightweight prediction architecture and a tailored guidance strategy that steers the generative process according to provided spatial targets. 
Crucially, as our method builds on the DiT's own learned features, our method naturally extends to video generation. This enables spatial manipulation (depth, pose, edge) and motion control (camera trajectories and motion dynamics),
under a single unified framework, requiring only a small dataset for training.

Compared to prior control methods, which typically address individual tasks or modalities, \name{} covers both spatial and motion control across image and video generation with a compact architecture and minimal training data. Since \name{} does not modify the base model, it can also be combined 
with adapter-based methods such as ControlNet for further improvement.
Our key contributions are the following:

\begin{itemize}[leftmargin=*,nosep]


    \item We introduce \name{}, a compact framework that enables  control 
    using DiT features extracted from a single transformer block.

    \item We extend our method to video generation, achieving spatial, camera, and motion manipulation without requiring large-scale video training overhead of traditional video control approaches.

    \item We provide comprehensive experiments demonstrating that our approach achieves competitive or improved guidance accuracy compared to feature-based and adapter-based approaches while using substantially fewer parameters, and demonstrating compatibility with adapter-based methods.

\end{itemize}


\section{Related Work}

\noindent\textbf{Controlling Diffusion Models.}
Diffusion models 
generate samples from the data distribution via 
gradual denoising,  
starting from a Gaussian noise sample and 
given a condition such as text. For text-to-image and text-to-video generation, it is common to use latent diffusion models \cite{rombach2021highresolution}, \ie, the  denoising is performed in the latent space defined by a pretrained variational autoencoder \cite{kingma2022autoencodingvariationalbayes}, which reduces the computational cost of generation. To train these models,  flow-matching \cite{lipman2023flowmatchinggenerativemodeling} is common. 


Beyond text-conditioned generation, many applications require 
controlling pretrained text-to-image diffusion models towards a given spatial target such as a depth-map, a pose-skeleton-map, or an edge-map. 
A widely adopted approach is to freeze the pretrained model and train an external adapter that directly interacts with the model's features.  ControlNet \cite{zhang2023adding} and T2I-Adapter \cite{mou2023t2i} train encoder branches that inject conditional signals into the denoising network, while ControlNet++ \cite{controlnet_plus_plus} further improves adherence through consistency feedback. More recent works achieve similar control with fewer parameters on DiT architectures. OminiControl \cite{ominicontrol, tan2025ominicontrol2} concatenates condition tokens into the DiT's attention stream, and EasyControl \cite{zhang2025easycontrol} introduces a lightweight LoRA-based injection module. These adapter-based methods achieve effective feedforward control, though they modify the base model's inputs or weights and are typically trained for a specific backbone.


Instead, we leverage learned diffusion features as control signals, building specifically upon ``Readout Guidance'' (RG) \cite{luo2024readoutguidancelearningcontrol}. RG trains an auxiliary model to predict spatial targets (\eg, depth, pose, edges) from a frozen diffusion model's features, using gradient updates on the latents to steer the generation process. We extend this framework to Diffusion Transformers (DiTs), offering new insights into DiT feature control and demonstrating its potential for video generation. 

\begin{table}[t]
\centering
\caption{\textbf{Comparison of DiT Control Approaches. }Overview of supported modalities, parameter overhead, training data requirements, and whether the method utilizes Implicit Model Conditioning (i.e., the diffusion model does not process the condition during its forward pass). For ControlNet, parameters for image and video are estimated from SD3-Medium and CogVideoX-5B respectively.}

\label{tab:method_comparison}
\setlength{\tabcolsep}{3pt}
\resizebox{\linewidth}{!}{
\begin{tabular}{l c c r c c}
\toprule
\textbf{Method} & \textbf{Image} & \textbf{Video} & \textbf{Params (M)} & \textbf{Data} & \textbf{\shortstack{Implicit Model\\Conditioning}} \\
\midrule
OminiControl        & \cmark & \xmark               & 14.5 & Small &  \xmark \\
ControlNet (Image)   & \cmark & \xmark               & 1487 & Small & \xmark \\
ControlNet (Video)   & \xmark & \cmark               & 2500 & Large & \xmark \\
RG                   & \cmark & \xmark               & 13.5 & Small & \cmark \\
DiTFlow              & \xmark & \cmark~(Motion only)  & 0    & None  & \xmark \\
Tora                 & \xmark & \cmark~(Motion only)  & 1100 & Large & \xmark \\
\midrule
ReaDiT (Ours)        & \cmark & \cmark               & 53   & Small & \cmark \\
\bottomrule
\end{tabular}
}
\vspace{-2em}
\end{table}


\noindent\textbf{Diffusion Transformer Insights.}
Several works aim to understand how large, pretrained diffusion models work internally and how their learned features can be readily applied to tasks, such as correspondence prediction \cite{tang2023emergent}, pose prediction \cite{liang2025sdposeexploitingdiffusionpriors}, depth prediction \cite{stracke2025cleandift}, and stylization \cite{Tumanyan_2023_CVPR}. However, many of these works study U-Net based diffusion models, which are no longer state-of-the-art for image and video generation. Meanwhile, while DiTs have achieved state-of-the-art for several modalities, an understanding of what the features of DiTs have learned is lacking. Our work sheds light in this direction as we investigate how to best leverage DiT features by developing a multi-resolution architecture inspired by \cite{ranftl2021visiontransformersdenseprediction} to predict spatial targets and use those predictions to control the generation.

\noindent\textbf{Video Control.}
As more text-to-video models are publicly available, interest in controlling the generated output increased. For motion control, prior works either train additional modules on top of video DiTs \cite{geng2024motionprompting, li2024imageconductorprecisioncontrol, zhang2025tora}, which is expensive, or employ separate LoRAs \cite{zhao2023motiondirector} for each motion, which is inefficient. 
Alternatively, training-free or optimization-based methods such as Frame Guidance \cite{jang2025frameguidance}, Time-to-Move \cite{singer2025timetomovetrainingfreemotioncontrolled}, and DiTFlow \cite{DITFlow} avoid additional training through denoising manipulation or attention flow optimization. However, these methods primarily address motion or camera control and are not uniformly applicable to multiple spatial tasks for both image and video.
Our work enables a more unified approach, applicable to both spatial and motion control. 

\noindent\textbf{Comparison with our approach.}
To summarize, adapter-based methods such as ControlNet and OminiControl achieve effective spatial control by feeding conditions directly into the diffusion model, but require modifying the base model's internal representation. Video control methods like Tora and DiTFlow address motion guidance but do not cover spatial tasks. Feature-based methods like RG avoid modifying the base model but do not necessarily extend to modern video models. Our method achieves unified spatial and motion control across both image and video generation, without modifying the base model or requiring large-scale training data. 
Table ~\ref{tab:method_comparison} summarizes this.

\section{\name{} Guidance}

\begin{figure*}[t]
    \centering
    \includegraphics[width=0.65\linewidth]{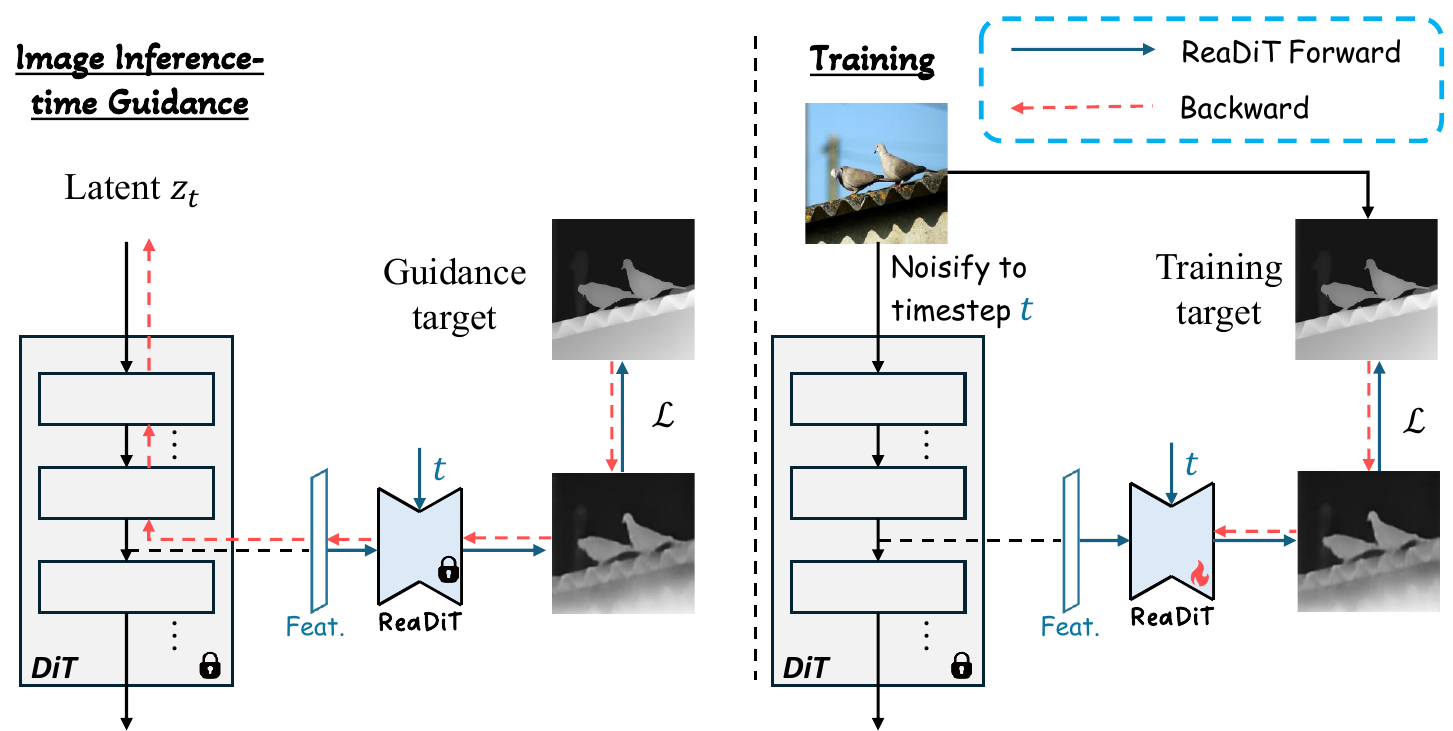}
    \caption{\textbf{Method Overview. }\textbf{Left:} \name{} takes a single intermediate feature from the frozen DiT and predicts spatial outputs, which are used to guide image generation by optimizing the latent $z_t$. \textbf{Right:} \name{} is trained to predict task-specific outputs from DiT features extracted by noising an input image to timestep $t$ and passing it through the frozen DiT.}
    \label{fig:method}
    \vspace{-3ex}
\end{figure*}


Our goal is to guide the diffusion process encoded in a pre-trained and frozen transformer-based image- or video-generation model (DiT) towards a given guidance target $y$. This is illustrated in Figure~\ref{fig:method} (left). In doing so, we want generated images or videos to remain diverse while simultaneously following the  given guidance target $y$. 

To achieve this, we develop a ``readout'' module $\hat y = g_\theta(\{f_t^l\}_{l\in L},t)$, which predicts the desired output $\hat y$ from a set of DiT features $\{f_t^l\}_{l\in L}$ extracted at diffusion time $t$ from DiT block $l\in L$ of a pre-trained DiT-based image- or video-generation model. We let $L$ denote the set of all DiT layers used for prediction in our ``readout'' module.  
We detail the architecture of our developed transformer-based readout module $g_\theta$ in Sec.~\ref{sec:module},  training in Sec.~\ref{sec:training}, and  use of the module for guidance in Sec.~\ref{sec:guidance}. We extend our method to guidance for video generation in Sec.~\ref{sec:video_guidance}.

\subsection{DiT Readout Model \name}
\label{sec:module}

\begin{figure*}[t]
    \centering
    \includegraphics[width=0.75\linewidth]{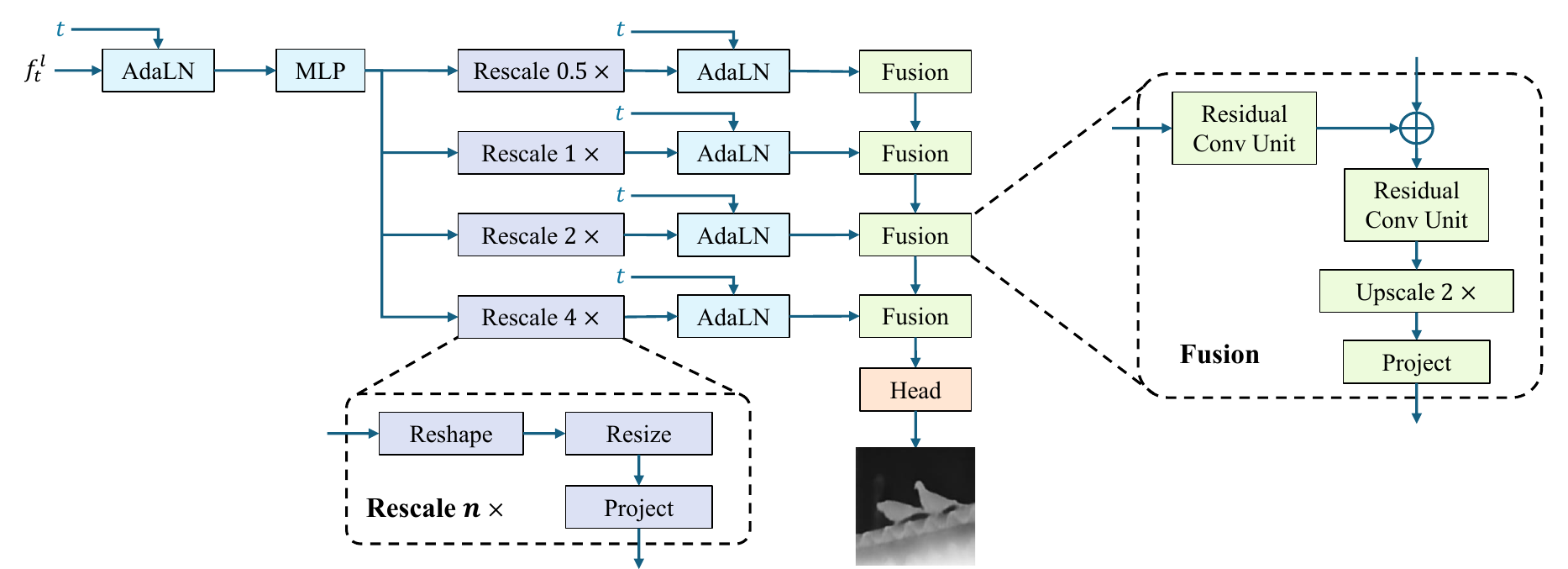}
    \caption{\textbf{Architecture of \name{}. }Input features are conditioned on $t$ and encoded into multi-resolution feature maps. The feature maps are progressively fused and upsampled to a final feature map, from which a task-specific head produces the spatial output. }
    \label{fig:readout_arch}
    \vspace{-3ex}
\end{figure*}

We highlight that the same architecture is used across all possible outputs, while only the training target differs. An overview of our  \name{} architecture is shown in Figure~\ref{fig:readout_arch}. Generally, the network $g_\theta(\{f_t^l\}_{l\in L},t)$ takes DiT features $f_t^l \in \mathbb{R}^{N \times D}$ $\forall l\in L$ and conditioning timestep $t$ as input. Here, $N$ denotes the number of tokens and $D$ denotes the feature dimension. 

Importantly, we find features extracted from DiT-based models to differ significantly from those obtained from U-Net-based models like SDXL. Concretely, we find U-Net-based features are much more diverse across layers, while DiT-based features are very similar across different DiT blocks. We suspect this is due to the nature of the DiT architecture: U-Net models are convolutional, \ie, each layer is more localized, whereas transformers process the output sequentially and therefore learn a global representation. 
%
%
Using this observation, we find that, as illustrated in Figure~\ref{fig:method}(left),  a  feature $f_t^l$ extracted from a single DiT block $l$ generally suffices for predicting compelling readouts $\hat y$. This finding has been corroborated by other works \cite{DiTF, yu2025repa}.
We hence use $|L| = 1$. 
We provide additional experiments and analysis in the supplementary material and discuss our readout model architecture next. 

We divide the network into three core components: encoder, rescale blocks, and fusion blocks.
The model's outputs are then passed into a task-specific head.


\noindent\textbf{Encoder. }Since DiT features are conditioned on $t$, their distribution varies across timesteps. To account for this, an AdaLN layer conditioned on $t$ is first applied to feature $f_t^l$. Note, $t$ produces scale $\gamma$ and shift $\beta$ to modulate the feature distribution. The modulated feature is then fed into a linear projection layer that reduces the channel dimension from $D$ to $D'$, producing a more compact feature. 

\noindent\textbf{Rescale blocks. }The compressed feature is reshaped from a token sequence   ($\mathbb{R}^{N \times D'}$) to a 2D spatial grid  ($\mathbb{R}^{h \times w \times D'}$). From this single-resolution grid, four parallel branches generate multi-scale features at $0.5\times$, $1\times$, $2\times$ and $4\times$ the spatial size through convolutional downsampling and upsampling respectively. Each branch applies another linear projection to map the features to a common channel dimension $D''$, resulting in multi-scale feature maps $\{r_i \in \mathbb{R}^{h_i \times w_i \times D''}\}_{i=1}^{4}$.

\noindent\textbf{Fusion blocks. }To prepare the multi-scale representations for decoding, an additional AdaLN conditioned on $t$ is applied to each feature map $r_i$. The maps are then progressively fused from the coarsest ($0.5h \times 0.5w \times D''$) to the finest ($4h \times 4w \times D''$) resolution. At each fusion stage, the coarser feature map is refined via spatial convolution and upsampled to double its spatial resolution. Then it is combined with the higher-resolution feature through a residual connection. A subsequent spatial convolution further refines the fused representation. This process repeats three more times to achieve the final resolution of $4h \times 4w \times D''$.

\noindent\textbf{Task-specific head. }The fused features are passed through a set of convolutional layers that map the latent representation to the target spatial map $\hat{y}$ with task-specific output channels.

In summary, the extracted DiT feature  $f_t^l$ is first modulated by $t$ and projected to  a multi-scale representation. These multi-scale representations are further modulated by $t$ and progressively fused to decode the final spatial map. We refer interested readers to the appendix for more details regarding the design of this architecture.



\subsection{Training}
\label{sec:training}

Next, we discuss training of our readout module $g_\theta$. This is illustrated in Figure~\ref{fig:method} (right). 
Our \name{} $g_\theta$ depends on features $f^l_t$ extracted from a frozen DiT at timestep $t$. For training, given an input image $x$, we add noise at a sampled timestep $t$ to its latent $z_0$ to obtain $z_t$. Intermediate features $f^l_t$ are extracted via a forward pass through the DiT, which we denote as $f_t^l(x)$. The training target $y$ is obtained by applying a state-of-the-art task-specific model to $x$, which we denote as $y(x)$. The general optimization applied across all tasks reads as follows:
\begin{equation}\label{eq:train}
    \min_\theta \mathbb{E}_{t\sim[t_{\min}, t_{\max}], x}\left[\mathcal{L}(g_\theta(f^l_t(x), t), y(x))\right].
\end{equation}
Here, ${\cal L}$ is a task-dependent loss which we discuss below. 

The choice of how $t$ is sampled significantly influences the training quality. While all timesteps encode partially useful information, only sampling from low-noise timesteps  ($t \rightarrow t_{\min}$) or over-sampling high-noise timesteps ($t \rightarrow t_{\max}$) is suboptimal. To balance the information across timesteps, 
we adopt a log-based sampling strategy as follows: $\log t \sim \mathcal{U}(\log t_{\min}, \log t_{\max})$. This strategy emphasizes lower-noise timesteps while still enabling the model to consider other timesteps.


The tasks we consider can be categorized into dense spatial prediction (depth, pose, edge) and sparse spatial correspondence prediction. 

\noindent\textbf{Dense spatial prediction tasks. }For dense spatial prediction tasks (depth, pose, edge), the target $y$ is a dense map such as a depth-map, a pose-skeleton-map, or an edge-map, produced by off-the-shelf state-of-the-art models. For these tasks, we use as the loss $\mathcal{L}$ the mean squared error (MSE) between prediction $\hat y = g_\theta(f_t^l,t)$ and target $y$.

\noindent\textbf{Sparse spatial correspondence prediction task. }The goal of the sparse spatial correspondence prediction task is to correctly predict the optical flow $y$ from a pair of frames. Given two related inputs $x_a$ and $x_b$, (\eg, two frames from a video clip), we extract the features $f^{l, a}_t$ and $f^{l, b}_t$ and pass them through \name{} $g_\theta$ to get two point descriptors $d_a, d_b \in \mathbb{R}^{N \times D''}$, where $N$ is the product of the point descriptors' height and width, and $D''$ stands for the point descriptors' channels. 
To find the corresponding relationships between the two frames, we compute a similarity matrix $S \in \mathbb{R}^{N \times N}$ via $S_{ij} = \frac{d_{a,i} \cdot d_{b, j}}{\|d_{a,i}\| \|d_{b,j}\|}$. We apply a row-wise softmax with a temperature parameter in order to get a probability distribution $P \in \mathbb{R}^{N \times N}$. 
For each pixel, we take the expected location 
as our final predicted value. To compute the expected location, let $G = \{(x, y) | x, y \in \{1, 2, \dots, N\}\} \in \mathbb{R}^{N \times 2}$ be a coordinate grid of all the coordinate pairs in the feature map. 
The expected location of a target point  is then $\hat{p}_{i} = \sum_{j = 1}^N P_{ij} G_j$, where $\hat{p}_{i} \in \mathbb{R}^2$. Given a set of source points, $p_{s}$, the predicted optical flow can be calculated as $\hat{y}_i = \hat{p}_{i}  - p_{s}$. For this task, the loss $\mathcal{L}$ is the L1 loss between the predicted optical flow $\hat{y}_i$ and the ground truth optical flow $y_i$. 

\begin{figure*}[t!]
    \centering
    \includegraphics[width=0.8\linewidth]{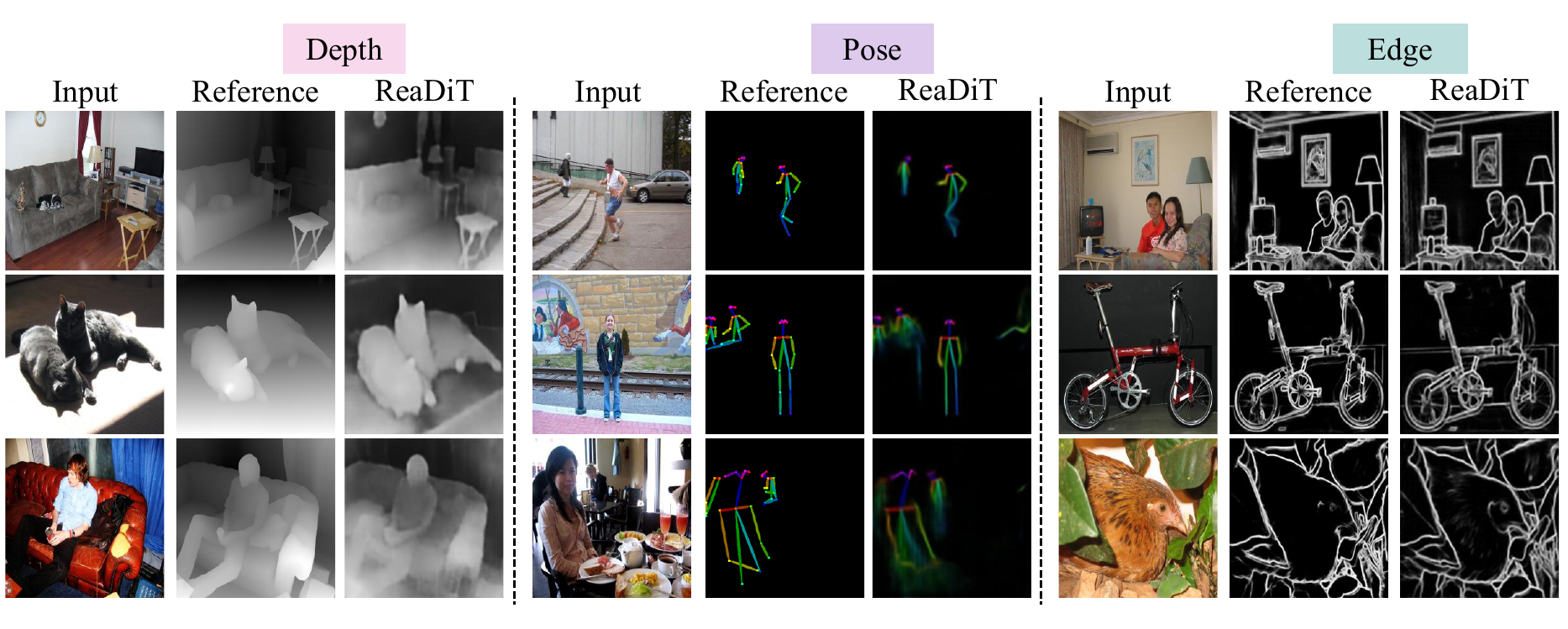}
    \caption{\textbf{Spatial Prediction Results. }We compare the prediction of \name{} to off-the-shelf models across depth, pose, and edge tasks. \name{} produces meaningful predictions from frozen DiT features, which can also be used to monitor spatial properties during the generation process. Additional examples are presented in the supplementary material. }
    \label{fig:prediction}
    \vspace{-3ex}
\end{figure*}

\subsection{Image Guidance}
\label{sec:guidance}

Via the trained \name{} $g_\theta$, we  predict the spatial output at any denoising timestep from the DiT features. We use the obtained prediction to guide the image generation process. Specifically, we update the latent $z_t$ at each denoising timestep $t$ such that the prediction $g_\theta(f_t^l, t)$ better matches a given guidance target $y$. The procedure is illustrated in  Figure~\ref{fig:method}(left). A general algorithm is provided in 
\ifarxiv
Algorithm~\ref{alg:guidance}.
\fi

At each denoising step, we compute the guidance loss between the prediction $\hat y = g_\theta(f_t^l,t)$ and target $y$, and update $z_t$ via gradient descent steps. 
Note, the update is repeated  $M$ times at each denoising timestep before the flow-matching step. During this process, the DiT and \name{} remain frozen, and only $z_t$ is updated. The guidance can be applied to all spatial tasks (depth, pose, edge) by changing the target $y$ and the corresponding trained \name{}. For insights regarding the refinements steps and learning rates, we refer readers to the appendix.

\ifarxiv
\begin{algorithm}[htbp]
    \footnotesize
    \SetAlgoLined
    \DontPrintSemicolon
    \SetNoFillComment
    \KwIn{Target $y$, denoising steps $T$, refinement steps $M$, learning rate $\eta$, frozen DiT $D$, \name{} $g_\theta$}
    \KwOut{Optimized latent $z_0$}
    
    \BlankLine
    Initialize $z_T \sim \mathcal{N}(0, I)$
    \For{$t = T$ \KwTo $0$}{
        \For{$i = 0$ \KwTo $M$}{
            $f_t^l = D(z_t^i, t)|_{block=l}$ \tcp*{Extract $l^{th}$ feat.}
            $\hat{y} = g_\theta(f_t^l, t)$ \tcp*{Predict spatial output}
            $\mathcal{L} = \mathcal{L}(y, \hat{y})$ \tcp*{Guidance loss}
            $z^{i + 1}_t \leftarrow z^{i}_t - \eta_t \nabla_{z_t} \mathcal{L} $
        }
        $v_t \leftarrow D(z_t^M, t)$ \tcp*{Predict velocity}
        $z_{t-1}^0 \leftarrow \text{Step}(v_t, t, z_t^M)$ \tcp*{Flow-match update}
    }
    \caption{\small Image Guidance via Latent Optimization}
    \label{alg:guidance}
\end{algorithm}
\vspace{-3ex}
\fi

\subsection{Video ReaDiT}
\label{sec:video_guidance}
Finally, we extend our method to DiT-based video generation models, as spatio-temporal features exhibit similar representation abilities to image DiTs. This includes spatial (depth, pose, and edge map) video guidance as well as motion-based (optical flow) video guidance.

\noindent\textbf{Video spatial guidance. }To train \name{} on video models, we follow the  training strategy for images: \name{} predicts a spatial output $\hat{y}=g_\theta(f_{t}^l, t)$ for a single frame. During inference, \name{} guides the generation toward a target sequence $Y = \{y_1, ..., y_K\}$ that specifies the desired spatial map for each frame. By applying the same latent optimization strategy described in Sec.~\ref{sec:guidance}, we minimize the loss $\mathcal{L}$ between the prediction $\hat{y}_k=g_\theta(f_{t, k}^l, t)$ and target $y_k$ summed across all frames $k$, and update the video latent. 
Note that some video architectures temporally downsample the latent space by a factor $s$, \ie, one latent frame corresponds to $s$ generated frames. In this case, the guidance loss $\mathcal{L}$ is computed between the $k$-th predicted output $\hat{y}_k$ with the $k \cdot s$-th target $y_{k \cdot s}$ from the target sequence. 

\noindent\textbf{Video motion guidance. }Using the sparse spatial correspondence training strategy described in Sec.~\ref{sec:training}, \name{} also learns to predict optical flow, which can be used to guide the video generation towards a reference optical flow. In this setting, we pass the features $f_{t, 1}^l$ and $f_{t, k}^l$ to \name{} to obtain point descriptors $d_1$ and $d_k$ for the first frame and the $k$-th frame. The prediction $\hat{y}_k$ represents the optical flow of the $k$-th frame relative to a reference frame (\eg, the first frame), calculated as $\hat{y}_k = \hat{p}_{k} - p_1$, where $\hat{p}_{k} $ is the expected value in frame $k$ corresponding to a source point $p_1$ in the reference frame. Similar to training, the guidance loss $\mathcal{L}$ is defined as the $\ell_1$ distance between the predicted flow $\hat{y}_k$ and a target optical flow $y_k$. Following the optimization process discussed in Sec.~\ref{sec:guidance}, the motion of the generated video can be guided to follow specific point trajectories.

\section{Experiments}

\subsection{Implementation Details}
\label{sec:implementation}
We implement our method on SD3-Medium~\cite{stablediffusion3} \& FLUX.1-dev ~\cite{flux2024} for images, and CogVideoX \cite{yang2025cogvideoxtexttovideodiffusionmodels} for video, both at 512 resolution. For image tasks, we use PascalVOC~\cite{everingham2010pascal} ($\sim$16K images) as the training dataset, with labels generated by off-the-shelf models: depth-maps from DepthAnythingV2~\cite{depth_anything_v2}, pose-skeleton-maps from OpenPose~\cite{cao2019openposerealtimemultiperson2d}, and edge-maps from HED~\cite{xie2015holisticallynestededgedetection}. For video spatial tasks, we train on single frames following the same protocol. For optical flow task, we use DAVIS videos~\cite{ponttuset20182017davischallengevideo} with point tracks from CoTracker3~\cite{karaev2024cotracker3simplerbetterpoint}. For both images and videos, training and guidance requires a single NVIDIA A40. 



\begin{figure*}[tb]
    \centering
    \includegraphics[width=0.70\linewidth]{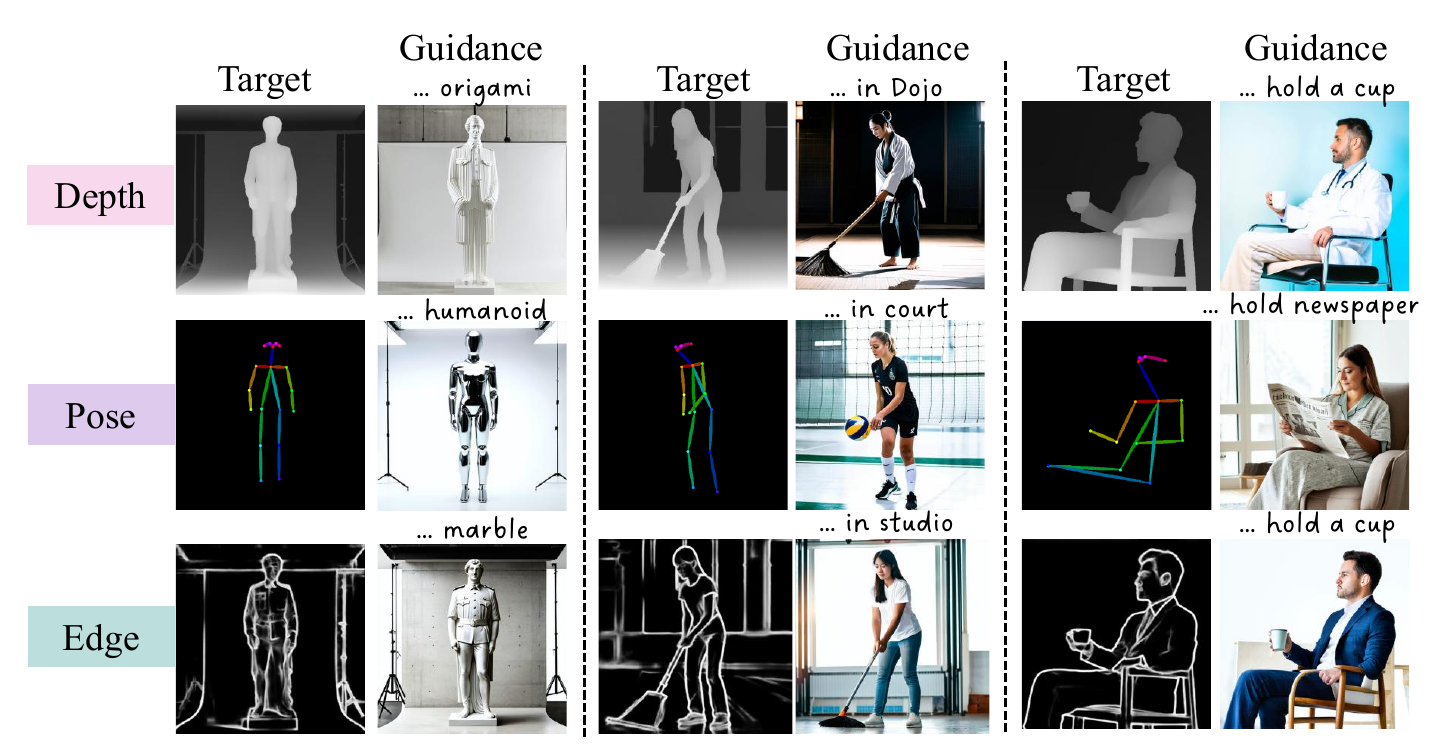}
    \caption{\textbf{Spatial Image Guidance. }Given a target spatial map, the trained \name{} guides the generation process to match the specific layout while maintaining image quality and adherence to the text prompt. Each column is generated from a text prompt following a shared format with varying suffixes, \eg, column1: ``A standing [suffix] in a studio'', where key suffixes are annotated in the figure. Full prompts and more results can be found in the supplementary material.}
    \label{fig:spatial_guidance}
    \vspace{-3ex}
\end{figure*}

\begin{figure*}[t]
    \centering
    \includegraphics[width=0.70\linewidth]{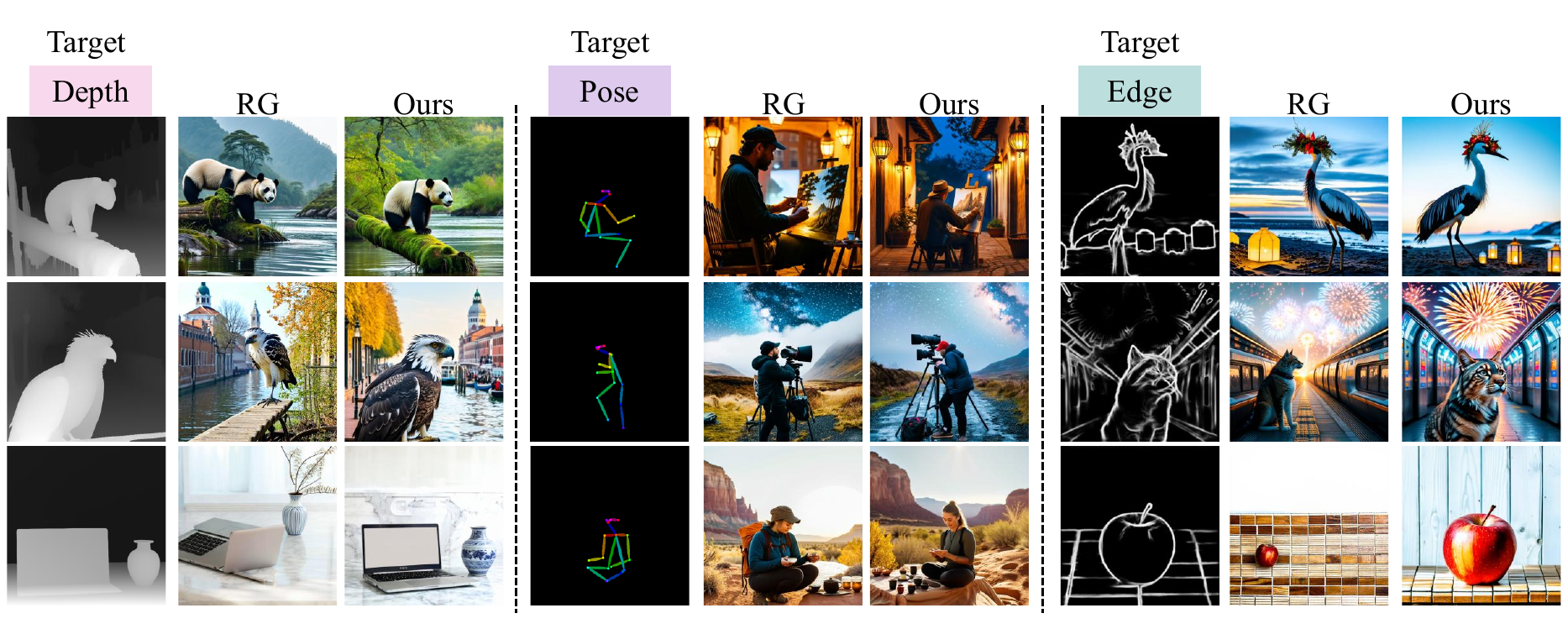}
    \vspace{-1.0em}
    \caption{\textbf{Qualitative Comparison. }Guidance of different spatial tasks from \name{} compared against RG~\cite{luo2024readoutguidancelearningcontrol}. Generation guided by \name{} follows the target more faithfully even under complex spatial maps. More examples are provided in the supplementary material.}
    \label{fig:sota_comparison}
    \vspace{-1em}
\end{figure*}

\subsection{Spatial Prediction Results}
We first evaluate the prediction quality of \name{} to validate that the network extracts meaningful spatial information from frozen DiT features. Figure~\ref{fig:prediction} shows qualitative prediction results across depth, pose and edge tasks. Despite being a lightweight network trained on a small dataset solely on DiT features, \name{} produces reasonable predictions that capture the overall spatial structure of the input, \eg, fine edge details and accurate human poses. This confirms that DiT features encode rich spatial information, and that \name{} can effectively extract it for task-specific use. Though \name{} does not always outperform task-specific models such as DepthAnythingv2 which is trained on over 62M images, it produces predictions sufficient for effective guidance.

\subsection{Spatial Image Guidance}
\label{sec:exp_image_guidance}

We evaluate image guidance on depth, pose, and edge tasks. Figure~\ref{fig:spatial_guidance} shows qualitative results across all three tasks. 
We consider two evaluation settings: (i) control maps from generated images, and (ii) control maps from real images. Setting (i) provides cleaner targets, while setting (ii) evaluates adherence under more challenging real-world conditions. 
Further details on the evaluation pipeline are provided in the .

\begin{table*}[t]
    \centering
    \setlength{\tabcolsep}{4.8pt}
    \caption{\textbf{Guidance Accuracy Comparison between \name{}, RG, and ControlNet.}
    We evaluate under two settings: control maps from generated images (Gen) and from real images (Real).
    Best model score per setting is \textbf{bolded}; second best is \underline{underlined}.
    \name{} outperforms RG across all tasks and metrics under both settings.
    With significantly fewer parameters, \name{} achieves comparable or better accuracy than ControlNet.
    Furthermore, \name{} is complementary to ControlNet, and combining the two further improves performance. We also compare against Flux-based OminiControl and implement our method on Flux.}
    \label{tab:image_quant}
    \small
    \renewcommand{\arraystretch}{1.2}
    \begin{tabular}{@{}lccccccccc@{}}
        \toprule
         &  
         & \multicolumn{2}{c}{\textbf{\colorbox[rgb]{0.97, 0.85, 0.93}{Depth}}} 
         & \multicolumn{4}{c}{\textbf{\colorbox[rgb]{0.87, 0.79, 0.93}{Pose}}} 
         & \multicolumn{2}{c}{\textbf{\colorbox[rgb]{0.74, 0.87, 0.87}{Edge}}} \\
        \cmidrule(lr){3-4} \cmidrule(lr){5-8} \cmidrule(lr){9-10}
         &  
         & \multicolumn{2}{c}{\textbf{RMSE ($\downarrow$)}} 
         & \multicolumn{2}{c}{\textbf{PCK@$\mathbf{0.2}$ ($\uparrow$)}} 
         & \multicolumn{2}{c}{\textbf{mAP ($\uparrow$)}} 
         & \multicolumn{2}{c}{\textbf{ODS ($\uparrow$)}} \\
        \cmidrule(lr){3-4} \cmidrule(lr){5-6} \cmidrule(lr){7-8} \cmidrule(lr){9-10}
        \textbf{Method} & \textbf{Params (M)} 
         & Gen & Real 
         & Gen & Real 
         & Gen & Real 
         & Gen & Real \\
        \midrule
        RG~\cite{luo2024readoutguidancelearningcontrol} 
            & 13.5 
            & 0.3704 & 0.4436 
            & 0.4881 & 0.3380 
            & 0.2739 & 0.1598 
            & \underline{0.5783} & 0.5733 \\
        ControlNet~\cite{zhang2023adding} 
            & 1487 
            & 0.2618 & \underline{0.3016} 
            & \underline{0.5012} & 0.4375 
            & 0.2946 & 0.2563 
            & 0.4908 & 0.5489 \\
        \name{} (Ours) 
            & 53 
            & \textbf{0.2492} & 0.3374
            & \textbf{0.5380} & \underline{0.4856} 
            & \textbf{0.3709} & \underline{0.3412} 
            & \textbf{0.6968} & \textbf{0.6644} \\
        ControlNet + \name{} (Ours) 
            & 1540 
            & \underline{0.2595} & \textbf{0.2961}
            & 0.4977 & \textbf{0.5386} 
            & \underline{0.3659} & \textbf{0.4067} 
            & 0.5688 & \underline{0.6192} \\
        \midrule
        OminiControl~\cite{ominicontrol}
            & 14.5
            & \textbf{0.0969} & \textbf{0.1290}
            & \underline{0.5649} & \underline{0.5143}
            & \underline{0.4818} & \underline{0.4657}
            & \textbf{0.7103} & \underline{0.6684} \\
        \name{} (Flux~\cite{flux2024}) (Ours)
            & 77.5
            & \underline{0.2734} & \underline{0.3427}
            & \textbf{0.6770} & \textbf{0.5758}
            & \textbf{0.5760} & \textbf{0.4711}
            & \underline{0.6727} & \textbf{0.7071} \\
        \bottomrule
    \end{tabular}
    \vspace{-4ex}
\end{table*}

\begin{figure*}
    \centering
    \vspace{2ex}
    \includegraphics[width=0.80\linewidth]{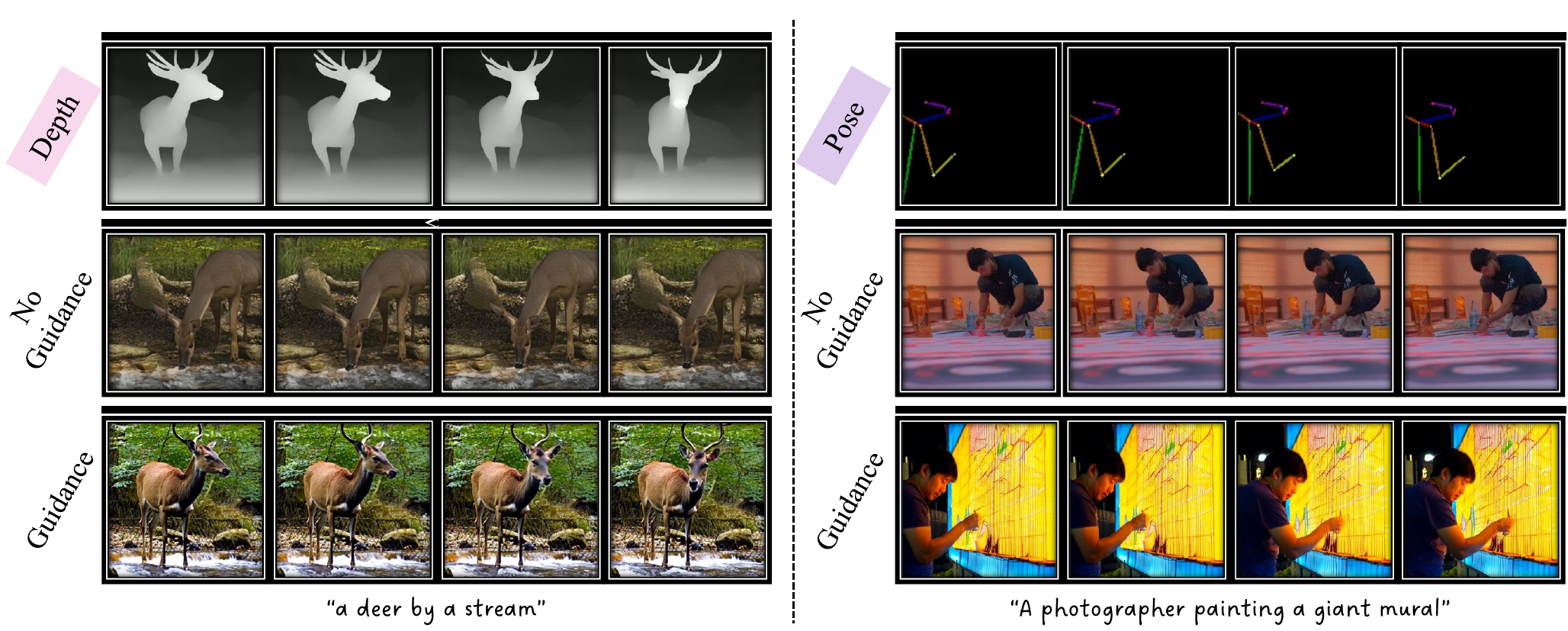}
    \caption{\textbf{Spatial Video Guidance. }Given a target spatial video input, ReaDiT guides the video generation process to match the specified spatial layout while maintaining visual quality and prompt adherence. Additional examples are shown in the supplementary material.}
    \label{fig:pose_video}
    \vspace{-3ex}
\end{figure*}

\begin{figure*}[t]
    \centering
    \includegraphics[width=0.7\linewidth]{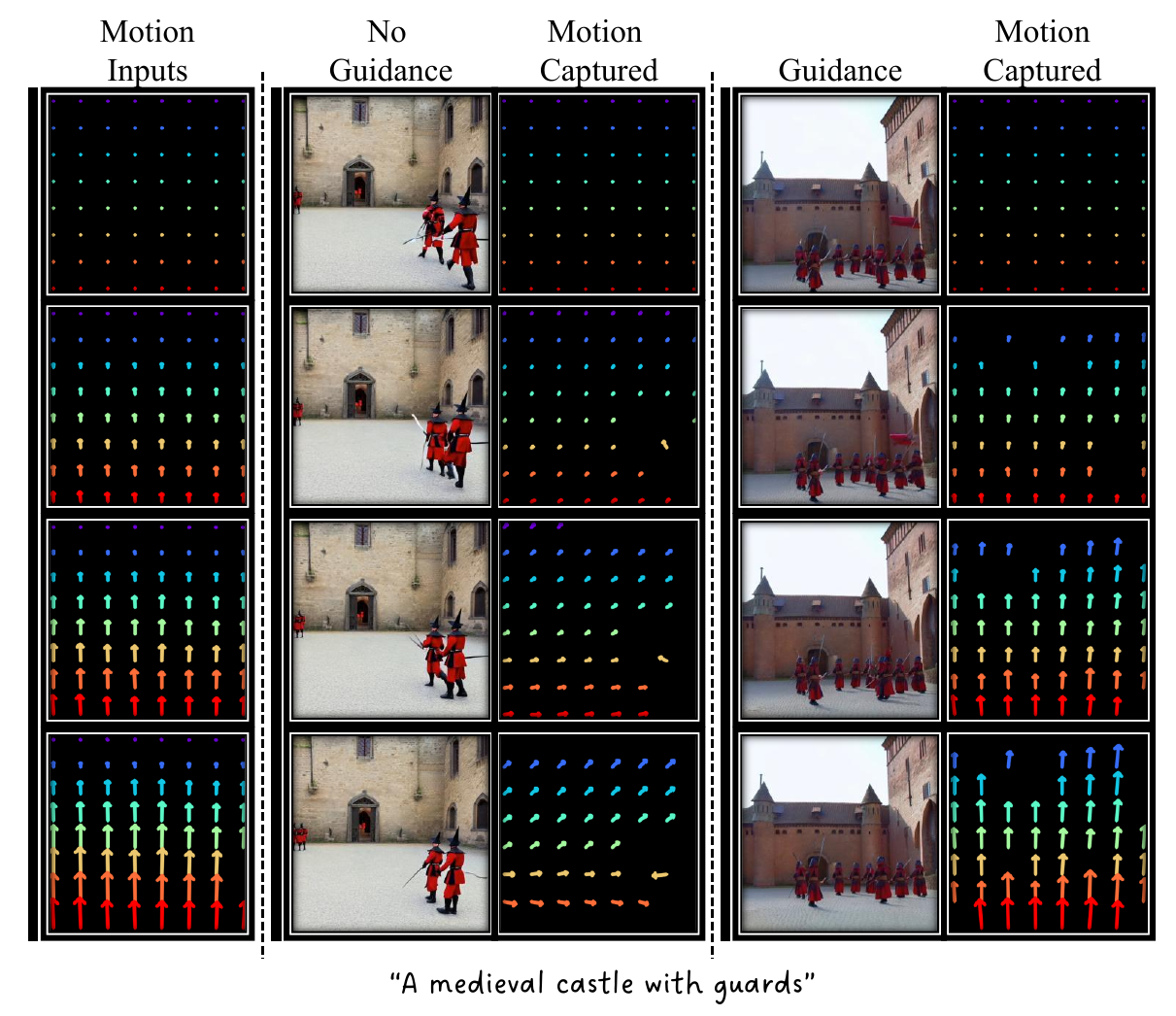}
    \vspace{-0.5em}
    \caption{\textbf{Motion Video Guidance Results. }From left to right: target motion inputs, unguided generation and its captured motion, guided generation and its captured motion. ReaDiT successfully guides video generation to follow the target motion input, as confirmed by the closer correspondence between the motion inputs and motion captured. Without guidance, the generated video fails to reproduce the intended motion.}
    \label{fig:motion}
    \vspace{-3ex}
\end{figure*}

We compare \name{} with RG~\cite{luo2024readoutguidancelearningcontrol}, which also uses diffusion features for control, but is designed for U-Net architectures and uses a convolution-based decoder for spatial tasks. We re-implement their method on SD3 using multiple DiT block features. Details of the re-implementation are provided in the supplementary material.

\noindent\textbf{Qualitative results. }Figure~\ref{fig:sota_comparison} shows a comparison across all three tasks. RG frequently fails to follow the input guidance signal, and when it does follow, the generated images may contain artifacts such as stretched and distorted objects. \name{} consistently follows the input target while producing natural and high-quality results across all tasks, improving upon RG. 

\noindent\textbf{Quantitative results. }Table~\ref{tab:image_quant} provides a quantitative comparison, assessing guidance accuracy for both evaluation settings. We apply the same off-the-shelf models on the generated images and compute the corresponding metrics against the input target. We report RMSE for depth, PCK@$0.2$~\cite{PCK_Andriluka_2014_CVPR} and mAP for pose, and ODS for edge, which are standard metrics for evaluating accuracy in these tasks. As shown in Table~\ref{tab:image_quant}, \name{} achieves better scores across all tasks and metrics under both settings, indicating more effective guidance that better adheres to the input target.


\noindent\textbf{Comparison with ControlNet and OminiControl. }Similarly, we compare our method to ControlNet and OminiControl. For a fair comparison, we train both methods 
on the same dataset as \name{} (PascalVOC) for all three tasks.
As shown in Table~\ref{tab:image_quant}, \name{} achieves comparable or better performance across most tasks, while using only $\sim$53M parameters compared to  ControlNet's $\sim$1487M parameters.
Furthermore, \name{} is compatible with adapter specific methods, such as ControlNet, and combining the two further improves performance across most tasks. For OminiControl, we implement our method on the same base model (Flux) and run the same evaluation suite. The results show that our method achieves comparable or better performance across most tasks. 

\ifarxiv
\noindent\textbf{Dual guidance. }Since \name{} guides generation via latent optimization, trained \name{} models for different modalities can be straightforwardly combined by backpropagating through all of their losses jointly. We demonstrate this by combining depth and pose guidance using control maps extracted from the same images. Compared to single-task guidance, dual guidance further improves adherence to both targets, as single-task guidance alone may not faithfully capture both the spatial structure and the human pose. For example, in Figure~\ref{fig:dual_consistent}, depth-only guidance does not recognize the statue as a human figure, while pose-only guidance does not reproduce the scene layout. Dual guidance resolves this by leveraging both signals simultaneously. This demonstrates the flexibility of \name{}. Further quantitative analysis of dual guidance is provided in the supplementary material.
 
Dual guidance can also be applied with inconsistent control maps, where the depth and pose targets are extracted from entirely different images. To handle the potential spatial conflict, our method automatically detects a pose mask and decouples the two guidance signals by applying the mask. Pose guidance is applied to the full image while depth guidance is restricted to the unmasked region. As shown in Figure \ref{fig:dual_inconsistent}, \name{} accurately follows both the depth layout and the pose target, producing high-quality generations even when the two control maps are spatially incompatible.

\begin{figure}[t]
    \centering
    \includegraphics[width=1.0\linewidth]{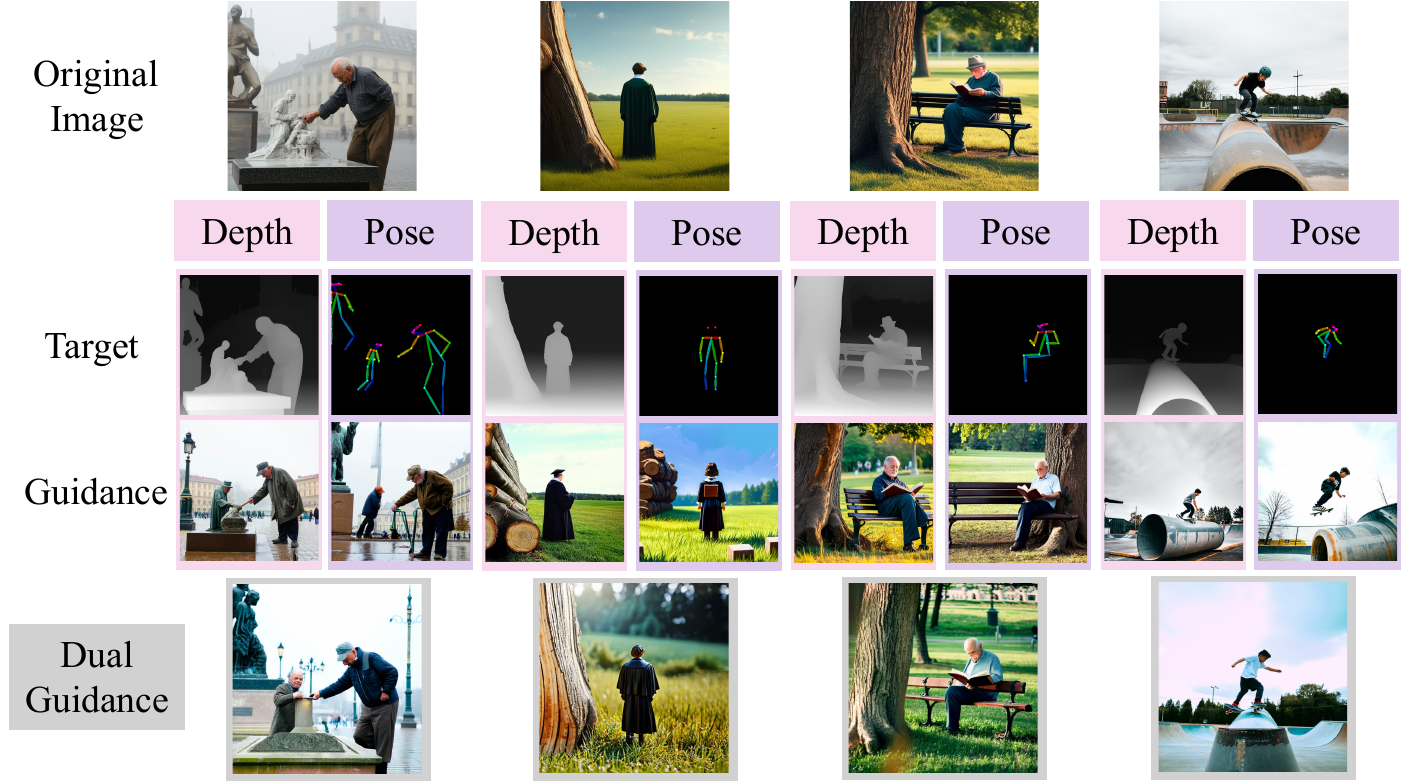}
    \caption{\textbf{Consistent Dual Guidance.} Depth and pose targets are extracted from the same source image. Compared to single-task guidance, dual guidance better captures both the scene geometry and the human pose.}
    \label{fig:dual_consistent}
    \vspace{-0.8em}
\end{figure}

\begin{figure}[hbt]
    \centering
    \includegraphics[width=1.0\linewidth]{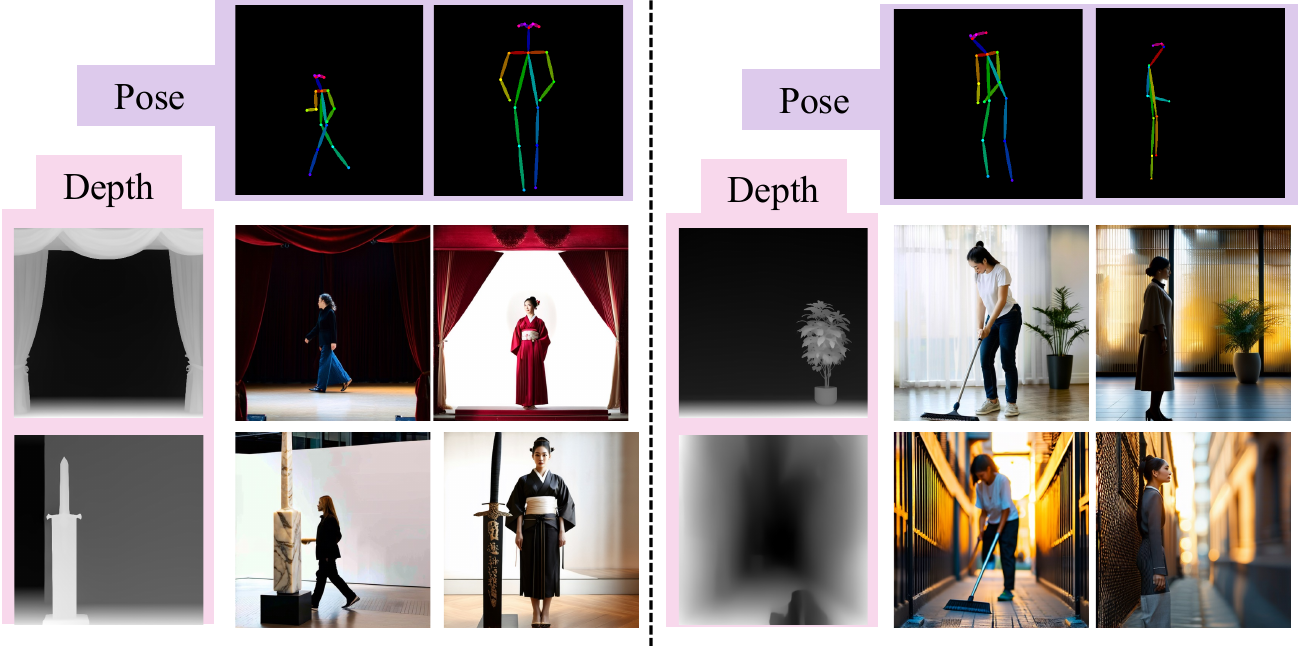}
    \caption{\textbf{Inconsistent Dual Guidance.} Depth and pose targets are extracted from different source images with potentially conflicting spatial layouts. Rows correspond to different depth targets, while columns to different pose targets. Despite the spatial mismatch, \name{} produces coherent, high-quality images that follow both the depth configuration and the pose.}
    \label{fig:dual_inconsistent}
    \vspace{-1em}
\end{figure}
\fi

\subsection{Video Guidance}

All video experiments use CogVideoX with temporal downsampling factor $s = 4$, following the training strategy in Sec.~\ref{sec:video_guidance} for each respective task. \name{} is trained separately for spatial guidance and motion (optical flow) guidance.

\noindent\textbf{Spatial guidance results.}
As shown in Figure~\ref{fig:pose_video}, \name{} adheres to the spatial video input while maintaining high video quality. Notably, despite not explicitly learning the temporal upsampling, \name{} still produces temporally consistent videos.

\noindent\textbf{Spatial guidance comparison. }Due to computation constraints, training a video ControlNet is infeasible.
Instead, we adapt RG on top of CogVideoX and follow the same evaluation setting as in Sec.~\ref{sec:exp_image_guidance} on a subset of videos. Further evaluation details are provided in the supplementary material.

As shown in Table~\ref{tab:video_quant}, \name{} shows higher adherence to spatial control than RG, both quantitatively and qualitatively (see supplementary material and supplementary material). RG’s performance degrades significantly under spatial video inputs, whereas \name{} successfully translates spatial video inputs into coherent videos. 



\noindent\textbf{Motion guidance results.}
As shown in Figure~\ref{fig:motion}, \name{} guides video generation to adhere to the input motion. Furthermore, \name{} supports various types of camera motions, such as panning right, left, up, down, as well as zooming in and out from a reference video or user-specified input, with additional results presented in the supplementary material.






\noindent\textbf{Motion guidance comparison. }
For motion guidance, we compare our work to ImageConductor~\cite{li2024imageconductorprecisioncontrol}, DiTFlow~\cite{DITFlow}, and Tora~\cite{zhang2025tora}. Since ImageConductor requires an input image, we use the first frame of our generated video as input for a fair comparison. Motion adherence is evaluated using endpoint error (EPE), defined as the $\ell_2$ distance between the predicted and input motion fields, represented as optical flow (see supplementary material for further details). 

As shown in Table~\ref{tab:user-study}, our method performs competitively against ImageConductor, but falls short of dedicated motion-specific methods. We attribute this gap primarily to the lack of temporal upsampling ability, which is crucial for capturing fast temporal dynamics. Addressing this remains a direction for future work.

\begin{table}[t]
\centering
\setlength{\tabcolsep}{3pt}
\caption{ \textbf{Spatial Video Guidance Comparison between \name{} and RG.}  Best model score is \textbf{bolded}. \name{} outperforms RG in all tasks and in all metrics demonstrating its superior performance and extendability to video guided generation. }
\renewcommand{\arraystretch}{1.2}
\scalebox{0.80}{%
    \begin{tabular}{@{}c c c cc c@{}}
    \toprule
     &  & \textbf{\colorbox[rgb]{0.97, 0.85, 0.93}{Depth}} & \multicolumn{2}{c}{\textbf{\colorbox[rgb]{0.87, 0.79, 0.93}{Pose}}}  & \textbf{\colorbox[rgb]{0.74, 0.87, 0.87}{Edge}} \\
    \cmidrule(lr){3-3} \cmidrule(lr){4-5} \cmidrule(lr){6-6}
    {\small \textbf{Method}} & {\small \textbf{Params (M)}}& {\small \textbf{RMSE ($\downarrow$)}} & {\small \textbf{PCK@$\mathbf{0.2}$ ($\uparrow$)}} & {\small \textbf{mAP ($\uparrow$)}} &  {\small \textbf{ODS ($\uparrow$)}} \\
    \midrule
    RG ~\cite{luo2024readoutguidancelearningcontrol} & 13.5 & 0.4610 & 0.1892 & 0.0496 & 0.5304 \\
    \name{} (Ours) & 53 & \textbf{0.4225} & \textbf{0.3651} & \textbf{0.1555} & \textbf{0.6101} \\
    \bottomrule
    \end{tabular}
}
\label{tab:video_quant}
\vspace{-2em}
\end{table}

\begin{table}[t]
    \centering
    \caption{\textbf{Motion Video Guidance Comparisons.} \name{} performs competitively against dedicated motion-control baselines. While our approach does not surpass the state of the art in quantitative metrics, it provides a highly versatile, architecture-agnostic alternative that can be seamlessly integrated with existing frameworks.}
    \label{tab:user-study}
    \small
    \setlength{\tabcolsep}{8pt}
    \begin{tabular}{@{}lrc@{}}
        \toprule
        \textbf{Method} & \textbf{Params (M)} & \textbf{EPE ($\downarrow$)} \\
        \midrule
        ImageConductor~\cite{li2024imageconductorprecisioncontrol} & 417 & 30.304 \\
        Tora~\cite{zhang2025tora} & 1100 & 16.100 \\
        DiTFlow~\cite{DITFlow} & 0 & 18.576 \\
        \midrule
        \name{} (Ours) & 53 & 29.644 \\
        \bottomrule
    \end{tabular}
    \vspace{-2em}
\end{table}

\section{Conclusion}
We introduced ReaDiT Guidance, a versatile method for controlling Diffusion Transformers via a single DiT feature. Our approach enables universal control across various vision task, including depth, pose and edge guidance for both image and video generation, as well as motion-guided video synthesis. 

\noindent\textbf{Limitations.}
As our method optimizes the latent during inference, it increases generation time compared to feedforward approaches. \Eg, guided image generation takes roughly 1 minute (vs.\ 30s unguided), and guided 16-frame video takes approximately 10 minutes (vs.\ 1.5 minutes unguided) on a single A40. 
Furthermore, motion guidance has difficulty with fast motion.  
Future work includes reducing inference cost through distillation and improving guidance for fast dynamics. 

\ifarxiv
    \vskip 1ex
    \noindent \textbf{Acknowledgments.}
    We would like to thank Aiyu Cui, Ozgur Kara, David Forsyth, Anand Bhattad, Vaibhav Vavilala, and Shivansh Patel for their helpful suggestions and comments. Finally, we thank Grace Luo, the first author of Readout Guidance, for helping with the implementation and answering specific questions of their prior work.

    This research used both the DeltaAI advanced computing and data resource, which is supported by the National Science Foundation (award OAC 2320345) and the State of Illinois, and the Delta advanced computing and data resource which is supported by the National Science Foundation (award OAC 2005572) and the State of Illinois. Delta and DeltaAI are joint efforts of the University of Illinois Urbana-Champaign and its National Center for Supercomputing Applications.
\fi 



{
    \small
    \bibliographystyle{ieeenat_fullname}
    \bibliography{vsbib}

\begin{thebibliography}{43}
\providecommand{\natexlab}[1]{#1}
\providecommand{\url}[1]{\texttt{#1}}
\expandafter\ifx\csname urlstyle\endcsname\relax
  \providecommand{\doi}[1]{doi: #1}\else
  \providecommand{\doi}{doi: \begingroup \urlstyle{rm}\Url}\fi

\bibitem[Andriluka et~al.(2014)Andriluka, Pishchulin, Gehler, and Schiele]{PCK_Andriluka_2014_CVPR}
Mykhaylo Andriluka, Leonid Pishchulin, Peter Gehler, and Bernt Schiele.
\newblock 2d human pose estimation: New benchmark and state of the art analysis.
\newblock In \emph{CVPR}, 2014.

\bibitem[Bai et~al.(2023)Bai, Bai, Yang, Wang, Tan, Wang, Lin, Zhou, and Zhou]{Qwen-VL}
Jinze Bai, Shuai Bai, Shusheng Yang, Shijie Wang, Sinan Tan, Peng Wang, Junyang Lin, Chang Zhou, and Jingren Zhou.
\newblock Qwen-vl: A versatile vision-language model for understanding, localization, text reading, and beyond.
\newblock \emph{arXiv preprint arXiv:2308.12966}, 2023.

\bibitem[Cao et~al.(2017)Cao, Hidalgo, Simon, Wei, and Sheikh]{cao2019openposerealtimemultiperson2d}
Zhe Cao, Gines Hidalgo, Tomas Simon, Shih-En Wei, and Yaser Sheikh.
\newblock Realtime multi-person 2d pose estimation using part affinity fields.
\newblock In \emph{CVPR}, 2017.

\bibitem[Duan et~al.(2025)Duan, Zhao, Yan, Li, Chen, Xu, Luo, Zhang, Gong, and Xia]{duan2025unic}
Lunhao Duan, Shanshan Zhao, Wenjun Yan, Yinglun Li, Qing-Guo Chen, Zhao Xu, Weihua Luo, Kaifu Zhang, Mingming Gong, and Gui-Song Xia.
\newblock Unic-adapter: Unified image-instruction adapter with multi-modal transformer for image generation.
\newblock In \emph{CVPR}, 2025.

\bibitem[Esser et~al.(2024)Esser, Kulal, Blattmann, Entezari, Müller, Saini, Levi, Lorenz, Sauer, Boesel, Podell, Dockhorn, English, Lacey, Goodwin, Marek, and Rombach]{stablediffusion3}
Patrick Esser, Sumith Kulal, Andreas Blattmann, Rahim Entezari, Jonas Müller, Harry Saini, Yam Levi, Dominik Lorenz, Axel Sauer, Frederic Boesel, Dustin Podell, Tim Dockhorn, Zion English, Kyle Lacey, Alex Goodwin, Yannik Marek, and Robin Rombach.
\newblock Scaling rectified flow transformers for high-resolution image synthesis.
\newblock In \emph{ICML}, 2024.

\bibitem[Everingham et~al.(2010)Everingham, Van~Gool, Williams, Winn, and Zisserman]{everingham2010pascal}
Mark Everingham, Luc Van~Gool, Christopher~KI Williams, John Winn, and Andrew Zisserman.
\newblock The pascal visual object classes (voc) challenge.
\newblock \emph{IJCV}, 2010.

\bibitem[Gan et~al.(2025)Gan, Tu, Chen, Chen, Li, Harandi, and Lin]{DiTF}
Chaofan Gan, Yuanpeng Tu, Xi Chen, Tieyuan Chen, Yuxi Li, Mehrtash Harandi, and Weiyao Lin.
\newblock Unleashing diffusion transformers for visual correspondence by modulating massive activations.
\newblock In \emph{NeurIPS}, 2025.

\bibitem[Geng et~al.(2025)Geng, Herrmann, Hur, Cole, Zhang, Pfaff, Lopez-Guevara, Doersch, Aytar, Rubinstein, Sun, Wang, Owens, and Sun]{geng2024motionprompting}
Daniel Geng, Charles Herrmann, Junhwa Hur, Forrester Cole, Serena Zhang, Tobias Pfaff, Tatiana Lopez-Guevara, Carl Doersch, Yusuf Aytar, Michael Rubinstein, Chen Sun, Oliver Wang, Andrew Owens, and Deqing Sun.
\newblock Motion prompting: Controlling video generation with motion trajectories.
\newblock In \emph{CVPR}, 2025.

\bibitem[Jang et~al.(2026)Jang, Ki, Jo, Yoon, Kim, Lin, and Hwang]{jang2025frameguidance}
Sangwon Jang, Taekyung Ki, Jaehyeong Jo, Jaehong Yoon, Soo~Ye Kim, Zhe Lin, and Sung~Ju Hwang.
\newblock Frame guidance: Training-free guidance for frame-level control in video diffusion models.
\newblock In \emph{ICLR}, 2026.

\bibitem[Karaev et~al.(2024)Karaev, Makarov, Wang, Neverova, Vedaldi, and Rupprecht]{karaev2024cotracker3simplerbetterpoint}
Nikita Karaev, Iurii Makarov, Jianyuan Wang, Natalia Neverova, Andrea Vedaldi, and Christian Rupprecht.
\newblock Cotracker3: Simpler and better point tracking by pseudo-labelling real videos, 2024.

\bibitem[Kingma and Welling(2014)]{kingma2022autoencodingvariationalbayes}
Diederik~P Kingma and Max Welling.
\newblock Auto-encoding variational bayes.
\newblock In \emph{ICLR}, 2014.

\bibitem[Labs(2024)]{flux2024}
Black~Forest Labs.
\newblock Flux.
\newblock \url{https://github.com/black-forest-labs/flux}, 2024.

\bibitem[Li et~al.(2024{\natexlab{a}})Li, Yang, Kuang, Wu, Wang, Xiao, and Chen]{controlnet_plus_plus}
Ming Li, Taojiannan Yang, Huafeng Kuang, Jie Wu, Zhaoning Wang, Xuefeng Xiao, and Chen Chen.
\newblock Controlnet{++}: Improving conditional controls with efficient consistency feedback.
\newblock In \emph{ECCV}, 2024{\natexlab{a}}.

\bibitem[Li et~al.(2024{\natexlab{b}})Li, Wang, Zhang, Wang, Yuan, Xie, Zou, and Shan]{li2024imageconductorprecisioncontrol}
Yaowei Li, Xintao Wang, Zhaoyang Zhang, Zhouxia Wang, Ziyang Yuan, Liangbin Xie, Yuexian Zou, and Ying Shan.
\newblock Image conductor: Precision control for interactive video synthesis.
\newblock In \emph{AAAI}, 2024{\natexlab{b}}.

\bibitem[Liang et~al.(2025)Liang, He, Wang, Liao, Zhang, Chen, and Yuan]{liang2025sdposeexploitingdiffusionpriors}
Shuang Liang, Jing He, Chuanmeizhi Wang, Lejun Liao, Guo Zhang, Yingcong Chen, and Yuan Yuan.
\newblock Sdpose: Exploiting diffusion priors for out-of-domain and robust pose estimation, 2025.

\bibitem[Lin et~al.(2025)Lin, Cho, Zala, and Bansal]{lin2025ctrl}
H Lin, J Cho, A Zala, and M Bansal.
\newblock Ctrl-adapter: An efficient and versatile framework for adapting diverse controls to any diffusion model.
\newblock In \emph{ICLR}, 2025.

\bibitem[Lin et~al.(2014)Lin, Maire, Belongie, Bourdev, Girshick, Hays, Perona, Ramanan, Zitnick, and Dollár]{lin2015microsoftcococommonobjects}
Tsung-Yi Lin, Michael Maire, Serge Belongie, Lubomir Bourdev, Ross Girshick, James Hays, Pietro Perona, Deva Ramanan, C.~Lawrence Zitnick, and Piotr Dollár.
\newblock Microsoft coco: Common objects in context.
\newblock In \emph{ECCV}, 2014.

\bibitem[Lipman et~al.(2023)Lipman, Chen, Ben-Hamu, Nickel, and Le]{lipman2023flowmatchinggenerativemodeling}
Yaron Lipman, Ricky T.~Q. Chen, Heli Ben-Hamu, Maximilian Nickel, and Matt Le.
\newblock Flow matching for generative modeling.
\newblock In \emph{ICLR}, 2023.

\bibitem[Luo et~al.(2024)Luo, Darrell, Wang, Goldman, and Holynski]{luo2024readoutguidancelearningcontrol}
Grace Luo, Trevor Darrell, Oliver Wang, Dan~B Goldman, and Aleksander Holynski.
\newblock Readout guidance: Learning control from diffusion features.
\newblock In \emph{CVPR}, 2024.

\bibitem[Mou et~al.(2023)Mou, Wang, Xie, Wu, Zhang, Qi, Shan, and Qie]{mou2023t2i}
Chong Mou, Xintao Wang, Liangbin Xie, Yanze Wu, Jian Zhang, Zhongang Qi, Ying Shan, and Xiaohu Qie.
\newblock T2i-adapter: Learning adapters to dig out more controllable ability for text-to-image diffusion models.
\newblock \emph{arXiv preprint arXiv:2302.08453}, 2023.

\bibitem[Peebles and Xie(2023)]{Peebles2023DiT}
William Peebles and Saining Xie.
\newblock Scalable diffusion models with transformers.
\newblock In \emph{ICCV}, 2023.

\bibitem[Podell et~al.(2024)Podell, English, Lacey, Blattmann, Dockhorn, Müller, Penna, and Rombach]{podell2023sdxlimprovinglatentdiffusion}
Dustin Podell, Zion English, Kyle Lacey, Andreas Blattmann, Tim Dockhorn, Jonas Müller, Joe Penna, and Robin Rombach.
\newblock Sdxl: Improving latent diffusion models for high-resolution image synthesis.
\newblock In \emph{ICLR}, 2024.

\bibitem[Pondaven et~al.(2025)Pondaven, Siarohin, Tulyakov, Torr, and Pizzati]{DITFlow}
Alexander Pondaven, Aliaksandr Siarohin, Sergey Tulyakov, Philip Torr, and Fabio Pizzati.
\newblock Video motion transfer with diffusion transformers.
\newblock In \emph{CVPR}, 2025.

\bibitem[Pont-Tuset et~al.(2018)Pont-Tuset, Perazzi, Caelles, Arbeláez, Sorkine-Hornung, and Gool]{ponttuset20182017davischallengevideo}
Jordi Pont-Tuset, Federico Perazzi, Sergi Caelles, Pablo Arbeláez, Alex Sorkine-Hornung, and Luc~Van Gool.
\newblock The 2017 davis challenge on video object segmentation, 2018.

\bibitem[Radford et~al.(2021)Radford, Kim, Hallacy, Ramesh, Goh, Agarwal, Sastry, Askell, Mishkin, Clark, et~al.]{radford2021clip}
Alec Radford, Jong~Wook Kim, Chris Hallacy, Aditya Ramesh, Gabriel Goh, Sandhini Agarwal, Girish Sastry, Amanda Askell, Pamela Mishkin, Jack Clark, et~al.
\newblock Learning transferable visual models from natural language supervision.
\newblock In \emph{ICML}, 2021.

\bibitem[Ranftl et~al.(2021)Ranftl, Bochkovskiy, and Koltun]{ranftl2021visiontransformersdenseprediction}
René Ranftl, Alexey Bochkovskiy, and Vladlen Koltun.
\newblock Vision transformers for dense prediction.
\newblock In \emph{ICCV}, 2021.

\bibitem[Rombach et~al.(2022)Rombach, Blattmann, Lorenz, Esser, and Ommer]{rombach2021highresolution}
Robin Rombach, Andreas Blattmann, Dominik Lorenz, Patrick Esser, and Björn Ommer.
\newblock High-resolution image synthesis with latent diffusion models.
\newblock In \emph{CVPR}, 2022.

\bibitem[Singer et~al.(2026)Singer, Rotstein, Mann, Kimmel, and Litany]{singer2025timetomovetrainingfreemotioncontrolled}
Assaf Singer, Noam Rotstein, Amir Mann, Ron Kimmel, and Or Litany.
\newblock Time-to-move: Training-free motion controlled video generation via dual-clock denoising, 2026.

\bibitem[Stracke et~al.(2025)Stracke, Baumann, Bauer, Fundel, and Ommer]{stracke2025cleandift}
Nick Stracke, Stefan~Andreas Baumann, Kolja Bauer, Frank Fundel, and Björn Ommer.
\newblock Cleandift: Diffusion features without noise.
\newblock In \emph{CVPR}, 2025.

\bibitem[Tan et~al.(2025{\natexlab{a}})Tan, Liu, Yang, Xue, and Wang]{ominicontrol}
Zhenxiong Tan, Songhua Liu, Xingyi Yang, Qiaochu Xue, and Xinchao Wang.
\newblock Ominicontrol: Minimal and universal control for diffusion transformer.
\newblock In \emph{ICCV}, 2025{\natexlab{a}}.

\bibitem[Tan et~al.(2025{\natexlab{b}})Tan, Xue, Yang, Liu, and Wang]{tan2025ominicontrol2}
Zhenxiong Tan, Qiaochu Xue, Xingyi Yang, Songhua Liu, and Xinchao Wang.
\newblock Ominicontrol2: Efficient conditioning for diffusion transformers.
\newblock \emph{arXiv preprint arXiv:2503.08280}, 2025{\natexlab{b}}.

\bibitem[Tang et~al.(2023)Tang, Jia, Wang, Phoo, and Hariharan]{tang2023emergent}
Luming Tang, Menglin Jia, Qianqian Wang, Cheng~Perng Phoo, and Bharath Hariharan.
\newblock Emergent correspondence from image diffusion.
\newblock In \emph{NeurIPS}, 2023.

\bibitem[Tumanyan et~al.(2023)Tumanyan, Geyer, Bagon, and Dekel]{Tumanyan_2023_CVPR}
Narek Tumanyan, Michal Geyer, Shai Bagon, and Tali Dekel.
\newblock Plug-and-play diffusion features for text-driven image-to-image translation.
\newblock In \emph{CVPR}, 2023.

\bibitem[Xie and Tu(2015)]{xie2015holisticallynestededgedetection}
Saining Xie and Zhuowen Tu.
\newblock Holistically-nested edge detection.
\newblock In \emph{ICCV}, 2015.

\bibitem[Xu et~al.(2025)Xu, He, Shan, and Chen]{xuctrlora}
Yifeng Xu, Zhenliang He, Shiguang Shan, and Xilin Chen.
\newblock Ctrlora: An extensible and efficient framework for controllable image generation.
\newblock In \emph{ICLR}, 2025.

\bibitem[Yang et~al.(2024)Yang, Kang, Huang, Zhao, Xu, Feng, and Zhao]{depth_anything_v2}
Lihe Yang, Bingyi Kang, Zilong Huang, Zhen Zhao, Xiaogang Xu, Jiashi Feng, and Hengshuang Zhao.
\newblock Depth anything v2.
\newblock In \emph{NeurIPS}, 2024.

\bibitem[Yang et~al.(2025)Yang, Teng, Zheng, Ding, Huang, Xu, Yang, Hong, Zhang, Feng, Yin, Zhang, Wang, Cheng, Xu, Gu, Dong, and Tang]{yang2025cogvideoxtexttovideodiffusionmodels}
Zhuoyi Yang, Jiayan Teng, Wendi Zheng, Ming Ding, Shiyu Huang, Jiazheng Xu, Yuanming Yang, Wenyi Hong, Xiaohan Zhang, Guanyu Feng, Da Yin, Yuxuan Zhang, Weihan Wang, Yean Cheng, Bin Xu, Xiaotao Gu, Yuxiao Dong, and Jie Tang.
\newblock Cogvideox: Text-to-video diffusion models with an expert transformer.
\newblock In \emph{ICLR}, 2025.

\bibitem[Yu et~al.(2025)Yu, Kwak, Jang, Jeong, Huang, Shin, and Xie]{yu2025repa}
Sihyun Yu, Sangkyung Kwak, Huiwon Jang, Jongheon Jeong, Jonathan Huang, Jinwoo Shin, and Saining Xie.
\newblock Representation alignment for generation: Training diffusion transformers is easier than you think.
\newblock In \emph{ICLR}, 2025.

\bibitem[Zhang et~al.(2023)Zhang, Rao, and Agrawala]{zhang2023adding}
Lvmin Zhang, Anyi Rao, and Maneesh Agrawala.
\newblock Adding conditional control to text-to-image diffusion models.
\newblock In \emph{ICCV}, 2023.

\bibitem[Zhang et~al.(2018)Zhang, Isola, Efros, Shechtman, and Wang]{zhang2018lpips}
Richard Zhang, Phillip Isola, Alexei~A Efros, Eli Shechtman, and Oliver Wang.
\newblock The unreasonable effectiveness of deep features as a perceptual metric.
\newblock In \emph{CVPR}, 2018.

\bibitem[Zhang et~al.(2025{\natexlab{a}})Zhang, Yuan, Song, Wang, and Liu]{zhang2025easycontrol}
Yuxuan Zhang, Yirui Yuan, Yiren Song, Haofan Wang, and Jiaming Liu.
\newblock Easycontrol: Adding efficient and flexible control for diffusion transformer.
\newblock In \emph{ICCV}, 2025{\natexlab{a}}.

\bibitem[Zhang et~al.(2025{\natexlab{b}})Zhang, Liao, Li, Dai, Qiu, Zhu, Qin, and Wang]{zhang2025tora}
Zhenghao Zhang, Junchao Liao, Menghao Li, Zuozhuo Dai, Bingxue Qiu, Siyu Zhu, Long Qin, and Weizhi Wang.
\newblock Tora: Trajectory-oriented diffusion transformer for video generation.
\newblock In \emph{CVPR}, 2025{\natexlab{b}}.

\bibitem[Zhao et~al.(2023)Zhao, Gu, Wu, Zhang, Liu, Wu, Keppo, and Shou]{zhao2023motiondirector}
Rui Zhao, Yuchao Gu, Jay~Zhangjie Wu, David~Junhao Zhang, Jiawei Liu, Weijia Wu, Jussi Keppo, and Mike~Zheng Shou.
\newblock Motiondirector: Motion customization of text-to-video diffusion models.
\newblock \emph{arXiv preprint arXiv:2310.08465}, 2023.

\end{thebibliography}
}



\clearpage
\resumetoc 
\renewcommand\thesection{\Alph{section}}
\renewcommand\thesubsection{\thesection.\arabic{subsection}}
\setcounter{section}{0}
\renewcommand\theHsection{supp.\Alph{section}}
\renewcommand\theHsubsection{supp.\thesection.\arabic{subsection}}
\maketitlesupplementary
\setcounter{tocdepth}{3}
\tableofcontents

\section{Architecture and Training Details}

\subsection{\name{} Architecture Details}



As discussed in Sec.~\ref{sec:module}, we design a time-conditioned \name{} architecture based on AdaLN. Since Diffusion Transformer (DiT) \cite{Peebles2023DiT} features are conditioned on timestep $t$, their distribution varies across diffusion timesteps. Additionally, prior work~\cite{DiTF} finds that AdaLN layers in DiT blocks introduce massive activations, \ie, extreme outlier activation values that negatively impact downstream spatial prediction tasks. To take into account both points, we apply AdaLN layers conditioned on $t$ to modulate the features, which suppresses outlier activations while preserving the timestep-dependent information encoded in the features.


For the task-specific head, when considering dense spatial prediction tasks, we use a small convolution network to map the fused latent representation to the target spatial map. When considering the sparse spatial correspondence prediction task, no head is needed as the predicted point descriptors are directly used to compute the optical flow, as described in Sec.~\ref{sec:training}.

\subsection{Training Details}
Table~\ref{tab:training_details} summarizes the training hyperparameters to train \name{} for all tasks. 
We extract features from a single block in each backbone: block 12 for SD3-Medium~\cite{stablediffusion3}, joint block 18 for FLUX.1 [dev]~\cite{flux2024}, and block 15 for CogVideoX~\cite{yang2025cogvideoxtexttovideodiffusionmodels}.

\begin{table*}[htb]
\centering
\caption{\small \textbf{Training Hyperparameters for ReaDiT.}}
\label{tab:training_details}
\renewcommand{\arraystretch}{1.2} 
\setlength{\tabcolsep}{6.8pt}
\small

\begin{tabular}{llccc}
\toprule
\textbf{Base Model} & \textbf{Spatial Task} & \textbf{Batch Size} & \textbf{Learning Rate} & \textbf{Epochs} \\
\midrule
SD3-Medium~\cite{stablediffusion3} & {Depth, Pose, Edge} & 8  & $1 \times 10^{-5}$ & 2  \\
Flux~\cite{flux2024} & {Depth, Pose, Edge} & 4  & $5 \times 10^{-5}$ & 2  \\
\midrule
\multirow{4}{*}{CogVideoX~\cite{yang2025cogvideoxtexttovideodiffusionmodels}} & {Depth} & \multirow{3}{*}{4}  & \multirow{3}{*}{$5 \times 10^{-5}$} & 3 \\
                            & {Pose}  &  & & 3 \\
                            & {Edge}  &  & & 2 \\ 
\cmidrule{2-5} 
                            & Motion & 1 & $1 \times 10^{-5}$ & 2 \\
\bottomrule
\end{tabular}

\end{table*}





\section{PCA Analysis}

Within the realm of diffusion, U-Net-based models differ significantly from DiT-based models. 
As our method is based on the intermediate features of the diffusion denoising network, we also study the internal representations of both architectures.
For this, we first use the Principal Component Analysis (PCA). We take two pretrained text-to-image models, the U-Net-based SDXL \cite{podell2023sdxlimprovinglatentdiffusion} and the DiT-based SD3-Medium (SD3) \cite{stablediffusion3}, and extract three feature maps from each of the 3 U-Net's up-blocks, and one feature map from SD3's last 9 sequential DiT blocks. For each output, we perform PCA and plot the first three principal components using RGB color images.

\begin{figure}[h]
    \centering
    \includegraphics[width=0.9\linewidth]{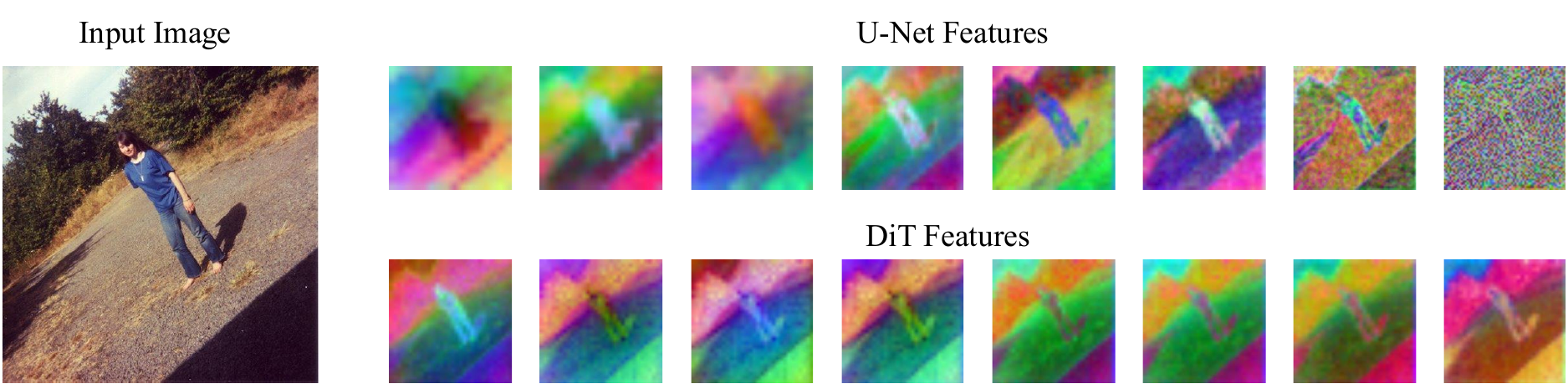}
    \caption{\textbf{U-Net (SDXL) Up-Block Features vs.\ DiT (SD3) Features from the last 9 Blocks.} U-Net features have much more variation across different layers compared to DiT features, which remain largely identical.
    Thus, while RG \cite{luo2024readoutguidancelearningcontrol} uses the $n$ last blocks, our \name{} builds upon the information extracted from a single block.}
    \label{fig:features}
\end{figure}

Figure~\ref{fig:features} shows the PCA visualizations of the extracted
features.
U-Net-based features are more diverse, capturing high and low level details of the image, 
while DiT features remain very similar 
across different blocks. 
To quantify this observation, we fit a linear mapping from one DiT block's features to each of the remaining blocks, obtaining an average (over the blocks) $R^2$ score of $0.7$, where $1.0$ indicates perfect linear dependence. This confirms that 
there is a significant degree of redundancy within the information provided by the different DiT blocks. 
We attribute this to the fundamental difference between the two
architectures. 
U-Net models are convolutional, forcing a hierarchical differentiation between the layers. In contrast, transformers
process the inputs at constant dimension and spatial resolution,
causing more similar token representations across blocks.

\begin{figure*}[htbp]
    \centering
    \includegraphics[width=1\linewidth]{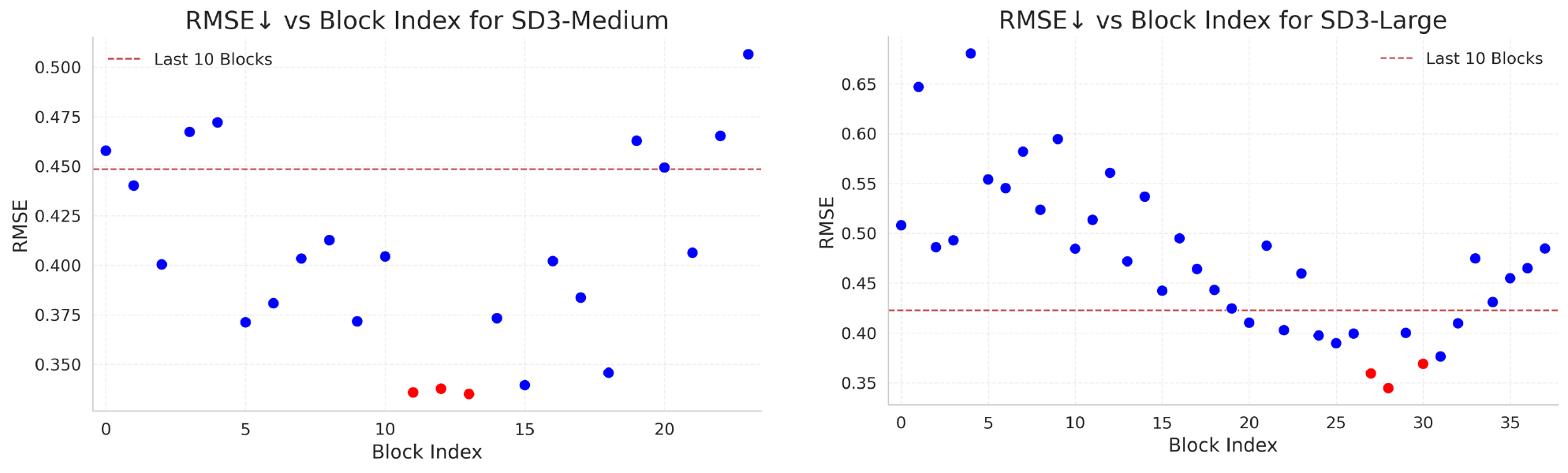}
    \caption{\textbf{RMSE ($\downarrow$) vs. Block Index for Depth Prediction Using Features from each DiT Block.} RMSE measures the accuracy of depth prediction. Results are shown for SD3-Medium (left) and SD3-Large (right). The red dots are the three lowest RMSE values obtained across all DiT blocks. The dotted red line is the RMSE using the last 10 blocks together. Using a single block to predict depth consistently outperforms using the last 10 blocks together.}
    \label{fig:RMSE}
\end{figure*}

This PCA analysis suggests that 
features from a single block may suffice for downstream prediction, which is also supported by observations in~\cite{yu2025repa}.
To verify this, we conduct an additional experiment: for
each $l$-th block, we extract feature $f_t^l$ and
train \name{} $g_\theta^l$ for depth prediction.
Each trained \name{} $g_\theta^l$ is then used to predict the depth map $\hat{y}$. We assess its quality via the RMSE $\sqrt{\frac{1}{M} \sum_{i=1}^{M} (y_i - \hat{y_i})^2}$, where $M$ denotes the total number of pixels and $y_i$ denotes the ground-truth depth map.
We also run the same experiment with 
features from the last ten blocks $f_t^{L}$ and train \name{} $g_\theta^L$, to assess whether 
multiple blocks are needed.



Figure~\ref{fig:RMSE} shows the RMSE of \name{} $g_\theta^l$ trained on features from the $l$-th DiT block, evaluated on both SD3-Medium and SD3-Large as base models. \name{} trained on a single DiT block's feature $f_t^l$ consistently achieves lower RMSE compared to \name{} trained on the combination of features from the last ten blocks, confirming that a single DiT block generally suffices for predicting compelling readouts $\hat{y}$.

Intuitively, using multiple blocks should do at least as well as using a single block since the multiple blocks contain at least as much information as a single block. However, in light of the multi-collinearity results and the earlier PCA plots, we believe that the redundant information between the DiT blocks may be inhibiting the performance of the model as the learned weights become more sensitive and unstable when using redundant inputs, thus causing worse generalization on unseen data. This suggests that a single block might be enough to accurately predict the spatial attribute of interest.

\begin{figure*}[t]
    \centering
    \includegraphics[width=0.9\linewidth]{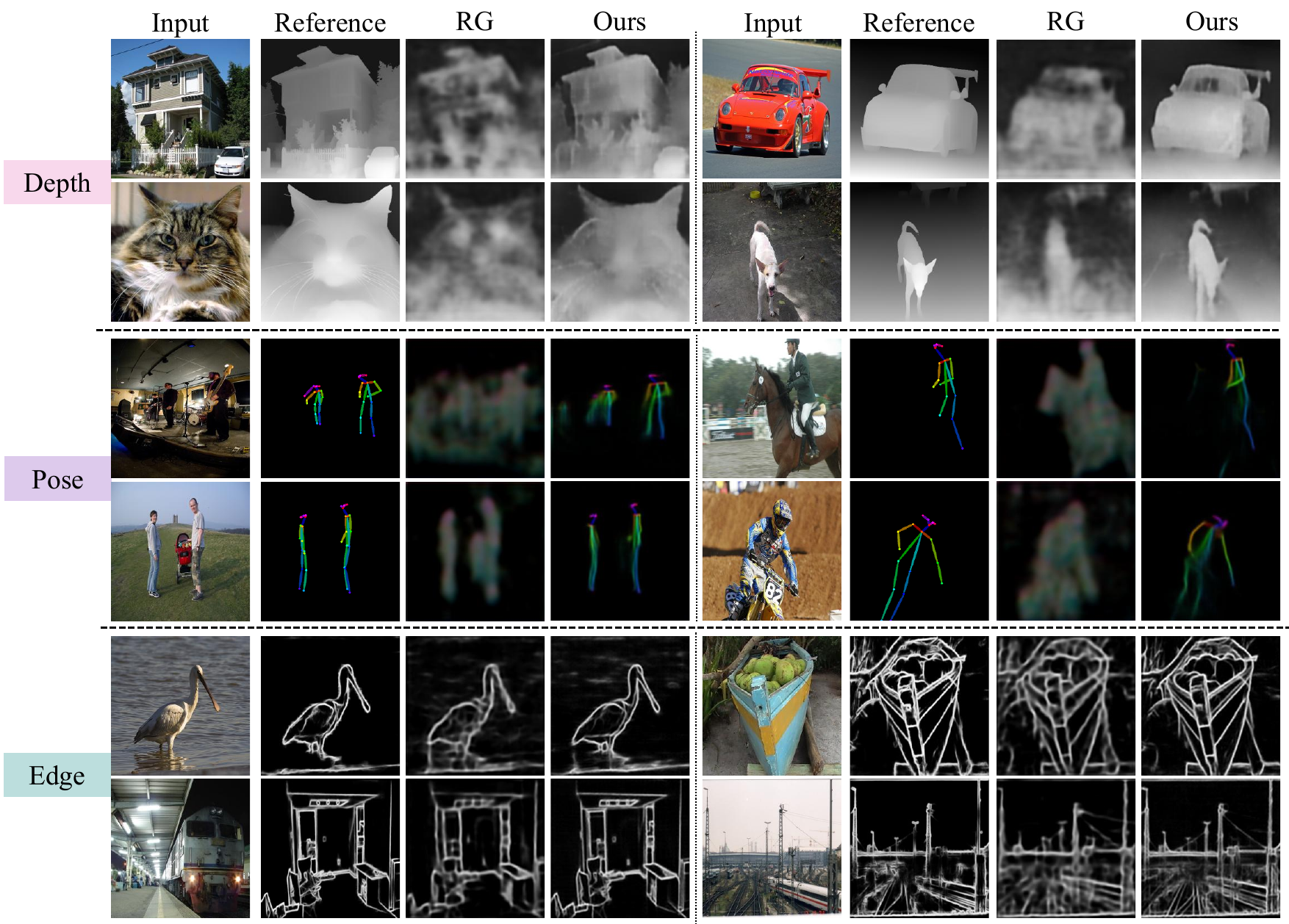}
    \caption{\textbf{Additional Spatial Prediction Results. }
    Qualitative comparison of spatial predictions from \name{} against RG~\cite{luo2024readoutguidancelearningcontrol} and off-the-shelf task-specific models across depth, pose, and edge tasks. \name{} consistently outperforms RG in prediction quality, demonstrating its superior ability to leverage DiT features for spatial understanding.
    }
    \label{fig:suppl_prediction}
    \vspace{-1em}
\end{figure*}

\begin{figure*}[t]
    \centering
    \includegraphics[width=0.8\linewidth]{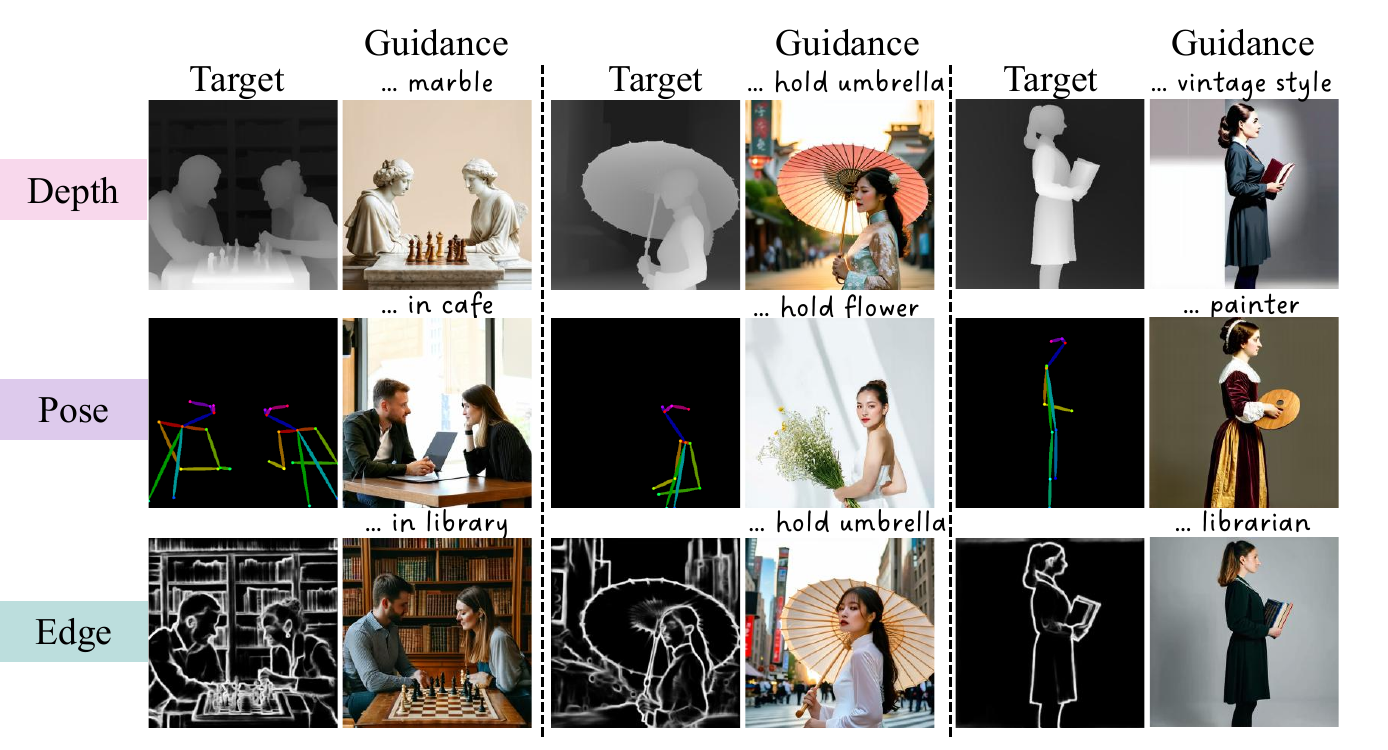}
    \caption{\textbf{Additional Spatial Image Guidance. }We present additional guided generation results given a target spatial map. The results again show that \name{} guides the outputs to matches the target spatial layout across diverse scenes and styles while preserving visual quality.
    }
    \label{fig:suppl_spatial_guidance}
\end{figure*}

\begin{figure*}[tb]
    \centering
    \includegraphics[width=0.7\linewidth]{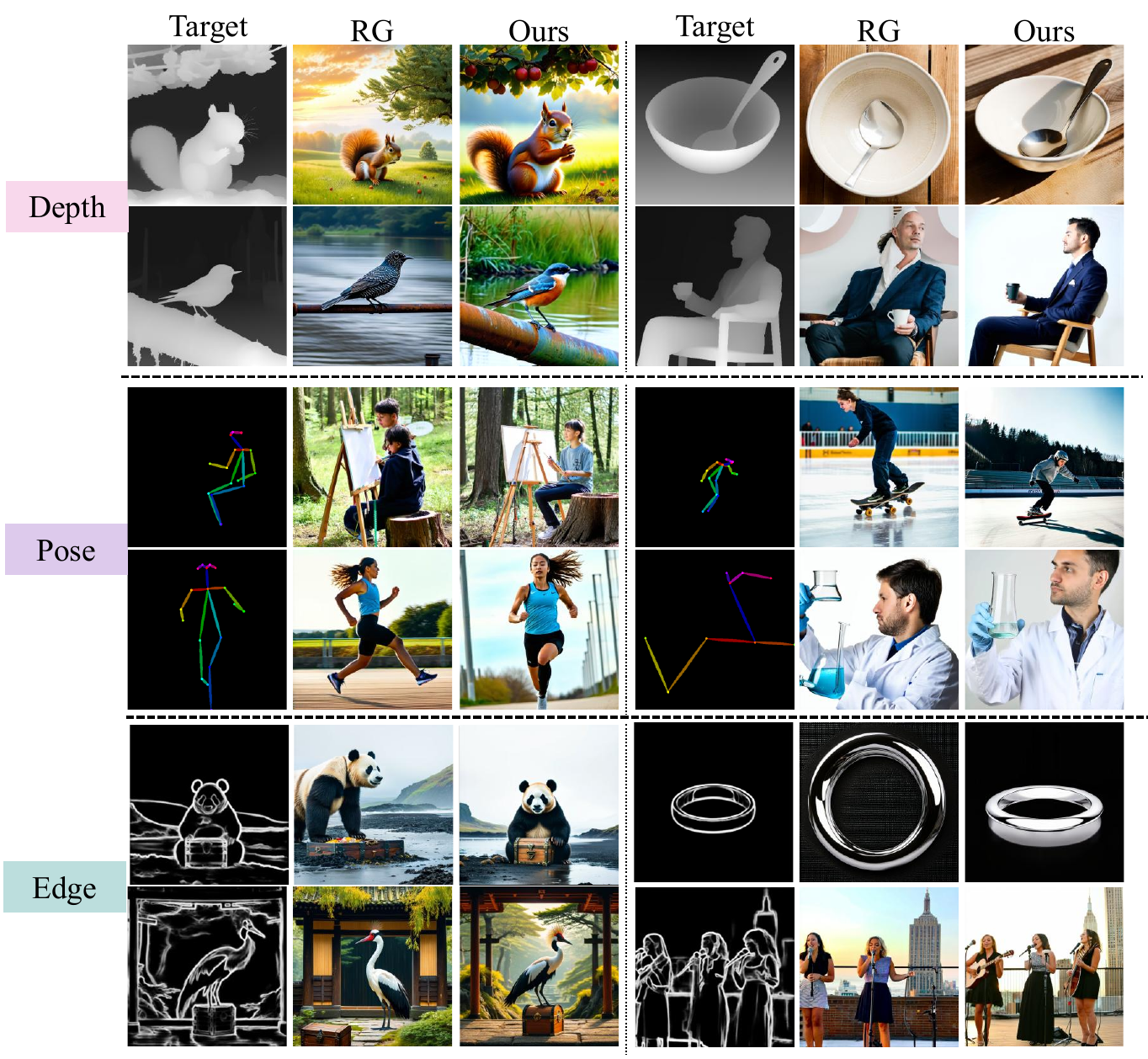}
    \caption{\textbf{Additional Qualitative Comparison on Image Guidance. }Additional group of guidance comparison between \name{} and RG~\cite{luo2024readoutguidancelearningcontrol}. \name{} consistently follows the target spatial map without artifacts observed in RG.}
    \label{fig:suppl_sota_comparison}
\end{figure*}

\section{Additional Results}

\subsection{Additional Results on Spatial Prediction}
\label{sec:suppl_prediction}

\name{} achieves competitive prediction performance compared to task-specific models despite being trained on a much smaller dataset. To further validate the quality of our learned features, we compare the prediction results with Readout Guidance (RG)~\cite{luo2024readoutguidancelearningcontrol} in Figure~\ref{fig:suppl_prediction}.

For depth prediction, RG produces noticeably noisier outputs that fail to recover clean subject boundaries and lack depth details. For pose prediction, RG captures only a rough map of the subject and struggles to produce a structured skeleton map, suggesting that RG struggles to extract semantically meaningful information from DiT features. Similarly, RG produces noisier edge maps. Across all three tasks, \name{} consistently produces cleaner and more structured predictions, which we attribute to the time-conditioned architecture and the log-based training strategy. Both enable a more effective extraction of spatial information from DiT features.

\begin{table*}[t]
    \centering
    \setlength{\tabcolsep}{4.8pt}
    \caption{\textbf{Guidance Accuracy Comparison between \name{}, RG, and RG with Our Guidance Strategy.}
    We evaluate under two settings: control maps from generated images (Gen) and from real images (Real).
    Best score per setting is \textbf{bolded}; second best is \underline{underlined}.
    Applying our guidance strategy to RG improves it on most tasks and metrics, but \name{} still outperforms both across nearly all settings.}
    \label{tab:rg_strategy}
    \small
    \renewcommand{\arraystretch}{1.2}
    \begin{tabular}{@{}lcccccccc@{}}
        \toprule
         & \multicolumn{2}{c}{\textbf{\colorbox[rgb]{0.97, 0.85, 0.93}{Depth}}} 
         & \multicolumn{4}{c}{\textbf{\colorbox[rgb]{0.87, 0.79, 0.93}{Pose}}} 
         & \multicolumn{2}{c}{\textbf{\colorbox[rgb]{0.74, 0.87, 0.87}{Edge}}} \\
        \cmidrule(lr){2-3} \cmidrule(lr){4-7} \cmidrule(lr){8-9}
         & \multicolumn{2}{c}{\textbf{RMSE ($\downarrow$)}} 
         & \multicolumn{2}{c}{\textbf{PCK@$\mathbf{0.2}$ ($\uparrow$)}} 
         & \multicolumn{2}{c}{\textbf{mAP ($\uparrow$)}} 
         & \multicolumn{2}{c}{\textbf{ODS ($\uparrow$)}} \\
        \cmidrule(lr){2-3} \cmidrule(lr){4-5} \cmidrule(lr){6-7} \cmidrule(lr){8-9}
        \textbf{Method} 
         & Gen & Real 
         & Gen & Real 
         & Gen & Real 
         & Gen & Real \\
        \midrule
        RG~\cite{luo2024readoutguidancelearningcontrol} 
            & \underline{0.3704} & 0.4436 
            & \underline{0.4881} & 0.3380 
            & 0.2739 & 0.1598 
            & 0.5783 & 0.5733 \\
        RG + Our Strategy 
            & 0.3720 & \underline{0.4257} 
            & 0.4846 & \underline{0.3595} 
            & \underline{0.2806} & \underline{0.1937} 
            & \textbf{0.6992} & \underline{0.6635} \\
        \name{} (Ours) 
            & \textbf{0.2492} & \textbf{0.3374}
            & \textbf{0.5380} & \textbf{0.4856} 
            & \textbf{0.3709} & \textbf{0.3412} 
            & \underline{0.6968} & \textbf{0.6644} \\
        \bottomrule
    \end{tabular}
\end{table*}

\subsection{Additional Results on Spatial Image Guidance}
\label{sec:suppl_image_guidance}

An advantage of \name{} is that it provides stable and consistent guidance across diverse scenes and styles. We provide additional qualitative results for spatial image guidance in Figure~\ref{fig:suppl_spatial_guidance}. The results further show that \name{} effectively guides the generation to adhere to the input spatial target while maintaining natural, high-quality outputs across various prompts and conditions.

We also present additional comparison results with RG in Table~\ref{tab:rg_strategy} and Figure~\ref{fig:suppl_sota_comparison}. RG is designed for U-Net backbones, whose decoder naturally provides features at multiple resolutions, whereas features from the DiT blocks are token sequences of the same length and width. We therefore adapt RG's readout architecture to accept DiT features. After extracting DiT features from different transformer blocks, we concatenate them along the channel dimension and reshape the token sequence into a 2D spatial grid. We then add several convolutional upsampling layers to turn this single-resolution grid into feature maps of multiple resolutions. The resulting maps are then passed to the RG network. At inference time, for a complete and fair comparison, we evaluate RG under two settings: (1) the strategy proposed in its original paper, and (2) our strategy, \ie, performing $M$ refinement steps at each denoising step, with $M$ set to the same value as in \name{} (Sec.~\ref{sec:guidance_details}).

As shown in Table~\ref{tab:rg_strategy}, RG improves under our guidance strategy on most tasks and metrics, yet \name{} still performs best on most of them. The advantage of \name{} therefore comes not only from the guidance strategy, but also from its readout architecture and training strategy.

Figure~\ref{fig:suppl_sota_comparison} qualitatively compares the two methods under the same guidance strategy. RG either struggles to follow the spatial target or forces adherence at the cost of visible artifacts, such as the distorted human in RG's depth guidance example. In contrast, \name{} achieves effective guidance without such degradation, producing artifact-free results that faithfully follow the input target.

\subsection{Prompt Influence on Spatial Image Guidance}

\begin{figure*}[t]
    \centering
    \includegraphics[width=0.9\linewidth]{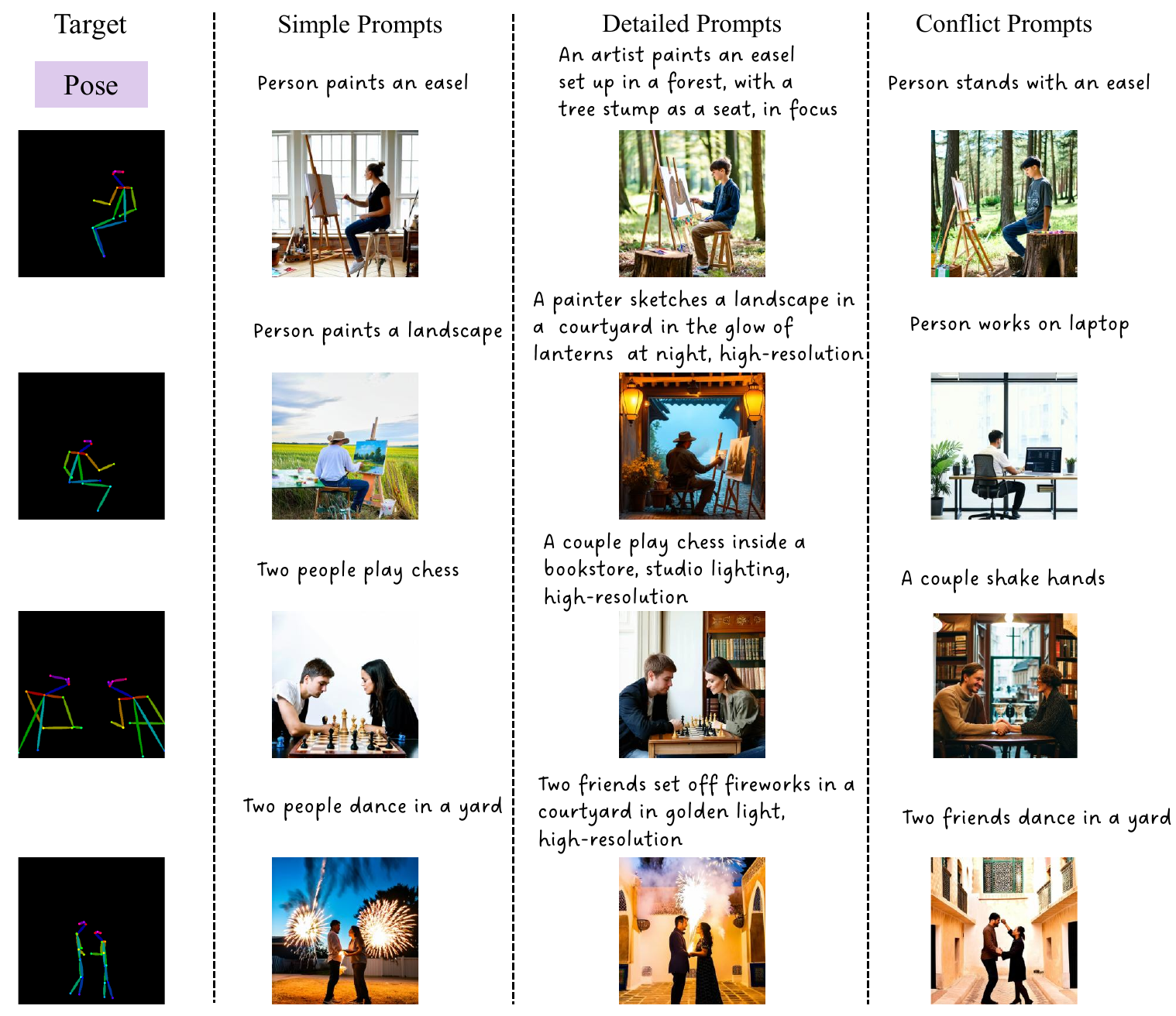}
    \caption{\textbf{Comparison of Spatial Image Guidance Results under Different Prompt Settings. }\name{} adheres to the input spatial target while respecting the text prompt, handling cases where the two partially conflict.}
    \label{fig:suppl_qualitativ_prompt}
\end{figure*}

As demonstrated in Sec.~\ref{sec:suppl_image_guidance}, \name{} performs well across various prompts. To further
illustrate how a prompt influences the guidance, we present a comparison of pose-guided generation with different prompts  in Figure~\ref{fig:suppl_qualitativ_prompt}. Although \name{} works well with simple prompts, more detailed prompts help generate richer details, such as the background and lighting. Furthermore, when the prompts partially conflict with the input target, \name{} successfully guides the conflicting parts to follow the prompt. For example, a couple shaking hands maintains the same body pose as the input target, while their hands follow the prompt description. This adherence to both the prompt and the input target further demonstrates the robustness of \name{}. 

\subsection{Additional Discussion on Dual Guidance}
\label{sec:suppl_dual_guidance}

We provide additional quantitative results and analyze how the individual guidance components influence the dual guidance outcome.

\paragraph{Quantitative results.}
Table~\ref{tab:supp_dual} compares the guidance accuracy of single-task guidance (depth only or pose only) against dual guidance (depth + pose jointly) under the consistent setting, where both control maps are extracted from the same source image. Dual guidance improves pose adherence compared to pose-only guidance, suggesting that the depth signal provides complementary geometric information that benefits pose prediction, particularly for incomplete or unusual poses. Depth accuracy remains comparable under dual guidance, which is expected. Pose guidance primarily affects the human region and has limited influence on the overall scene geometry as long as the guided output does not deviate substantially from the depth target.

\begin{table}[t]
\centering
\setlength{\tabcolsep}{5pt}
\caption{\textbf{Dual Guidance Quantitative Results (Consistent).}
Comparison of single-task and dual-task guidance accuracy. Dual guidance achieves comparable or better adherence to both targets, indicating that the two guidance signals are complementary.
}
\renewcommand{\arraystretch}{1}
\scalebox{0.82}{%
    \begin{tabular}{c cc cc cc}
    \toprule
     & \multicolumn{2}{c}{\textbf{\colorbox[rgb]{0.97, 0.85, 0.93}{Depth}}} 
     & \multicolumn{4}{c}{\textbf{\colorbox[rgb]{0.87, 0.79, 0.93}{Pose}}} \\
    \cmidrule(lr){2-3} \cmidrule(lr){4-7}
     & \multicolumn{2}{c}{\small \textbf{RMSE ($\downarrow$)}} 
     & \multicolumn{2}{c}{\small \textbf{PCK@$\mathbf{0.2}$ ($\uparrow$)}} 
     & \multicolumn{2}{c}{\small \textbf{mAP ($\uparrow$)}} \\
    \cmidrule(lr){2-3} \cmidrule(lr){4-5} \cmidrule(lr){6-7}
    {\small \textbf{Guidance}} 
     & {\small Gen} & {\small Real} 
     & {\small Gen} & {\small Real} 
     & {\small Gen} & {\small Real} \\
    \midrule
    Depth only 
        & 0.3234 & 0.3658 
        & -- & -- 
        & -- & -- \\
    Pose only 
        & -- & -- 
        & 0.3709 & 0.3412 
        & 0.5380 & 0.4856 \\
    \midrule
    Depth + Pose 
        & \textbf{0.3232} & \textbf{0.3534} 
        & \textbf{0.5773} & \textbf{0.3589} 
        & \textbf{0.6712} & \textbf{0.5017} \\
    \bottomrule
    \end{tabular}%
}
\label{tab:supp_dual}
\vspace{-1.5ex}
\end{table}


\begin{table*}[htbp]
\centering
\caption{\textbf{Influence of Guidance Ratio on Dual Guidance.}
We vary the ratio between depth and pose guidance losses. A balanced ratio yields the best overall performance across both tasks.
}
\label{tab:supp_factor}
\renewcommand{\arraystretch}{1.2}
\begin{subtable}[t]{0.48\textwidth}
\centering
\caption{Fix depth $= 4$, vary pose.}
\label{tab:supp_factor_pose}
\scalebox{0.92}{%
    \begin{tabular}{c c c c}
        \toprule
        {\small \textbf{Depth:Pose}} 
         & \multicolumn{1}{c}{\textbf{\colorbox[rgb]{0.97, 0.85, 0.93}{Depth}}}
         & \multicolumn{2}{c}{\textbf{\colorbox[rgb]{0.87, 0.79, 0.93}{Pose}}} \\
        \cmidrule(lr){2-2} \cmidrule(lr){3-4}
        {\small \textbf{Ratio}}
         & {\small \textbf{RMSE ($\downarrow$)}} 
         & {\small \textbf{mAP ($\uparrow$)}} 
         & {\small \textbf{PCK ($\uparrow$)}} \\
        \midrule
        4:1 \,\small{(4.0)}  & \textbf{0.2501} & 0.7706 & 0.8305 \\
        4:3 \,\small{(1.3)}  & 0.2724 & \textbf{0.8976} & \textbf{0.9546} \\
        4:8 \,\small{(0.5)}  & 0.2874 & 0.8602 & 0.9108 \\
        4:12 \,\small{(0.3)} & 0.3089 & 0.8298 & 0.9054 \\
        \bottomrule
    \end{tabular}%
}
\end{subtable}
\hfill
\begin{subtable}[t]{0.48\textwidth}
\centering
\caption{Fix pose $= 8$, vary depth.}
\label{tab:supp_factor_depth}
\scalebox{0.92}{%
    \begin{tabular}{c c c c}
        \toprule
        {\small \textbf{Depth:Pose}} 
         & \multicolumn{1}{c}{\textbf{\colorbox[rgb]{0.97, 0.85, 0.93}{Depth}}}
         & \multicolumn{2}{c}{\textbf{\colorbox[rgb]{0.87, 0.79, 0.93}{Pose}}} \\
        \cmidrule(lr){2-2} \cmidrule(lr){3-4}
        {\small \textbf{Ratio}}
         & {\small \textbf{RMSE ($\downarrow$)}} 
         & {\small \textbf{mAP ($\uparrow$)}} 
         & {\small \textbf{PCK ($\uparrow$)}} \\
        \midrule
        1:8 \,\small{(0.1)}   & 0.3377 & 0.8229 & 0.9099 \\
        4:8 \,\small{(0.5)}   & 0.2874 & 0.8602 & 0.9108 \\
        10:8 \,\small{(1.3)}  & 0.2744 & \textbf{0.8975} & \textbf{0.9495} \\
        16:8 \,\small{(2.0)}  & \textbf{0.2601} & 0.7979 & 0.8602 \\
        \bottomrule
    \end{tabular}%
}
\end{subtable}
\vspace{-2ex}
\end{table*}

\begin{figure*}[htbp]
    \centering
    \includegraphics[width=1\linewidth]{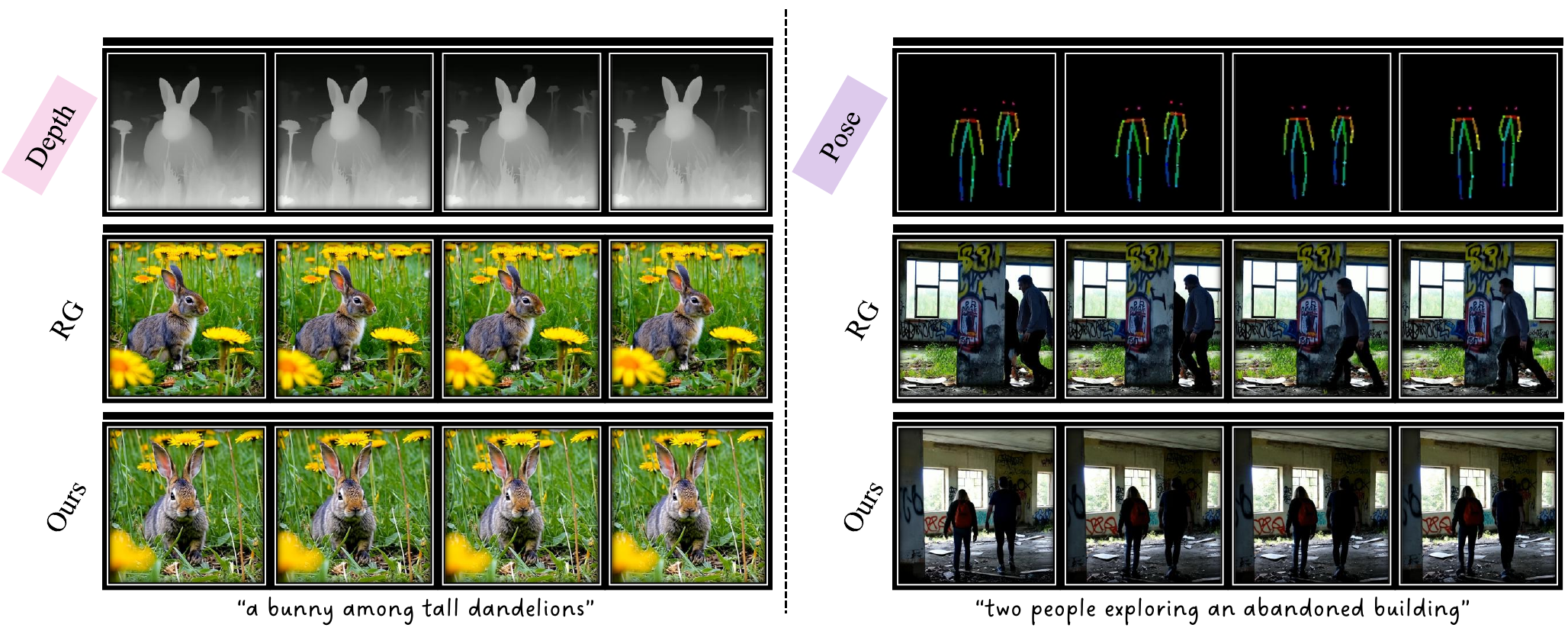 }
    \caption{\textbf{Qualitative Comparison for Spatial Video Guidance. }Compared to RG \cite{luo2024readoutguidancelearningcontrol}, ReaDiT demonstrates stronger spatial control across both depth and pose conditions, while generating consistent video sequences.}
    \label{fig:video_sp_compare}
    \vspace{-1em}
\end{figure*}

\paragraph{Influence of guidance weight ratio. }To better understand how each guidance signal contributes to dual guidance, we analyze the effect of varying the relative weight between depth and pose losses. Specifically, we restrict this analysis to a fixed subset of evaluation images in which both control signals are well posed, \ie, the human figure is fully visible so that pose guidance is not degenerate. On this subset we fix the weight of one task and sweep the weight of the other, measuring adherence to both targets. Table~\ref{tab:supp_factor_pose} shows the effect of varying the pose weight while keeping the depth weight fixed, and Table~\ref{tab:supp_factor_depth} shows the effect of varying the depth weight while keeping the pose weight fixed.

Across both sweeps, we observe a consistent tradeoff: increasing the weight of one task improves its adherence at the cost of the other. However, when the depth and pose weights are roughly balanced (ratio $\sim 1$), the model achieves the best overall performance across both tasks. Extreme ratios degrade one of the two targets noticeably. \eg, a high pose weight relative to depth yields the best pose metrics but weakens depth adherence, while a high depth weight relative to pose achieves the lowest depth RMSE but reduces pose accuracy. These results suggest that the two guidance signals compete for modulating latent, and a balanced ratio is necessary to maintain strong adherence to both targets simultaneously.

\subsection{Additional Results on Video Guidance}
\label{sec:suppl_video_guidance}

\begin{figure*}[t]
    \centering
    \includegraphics[width=0.9\linewidth]{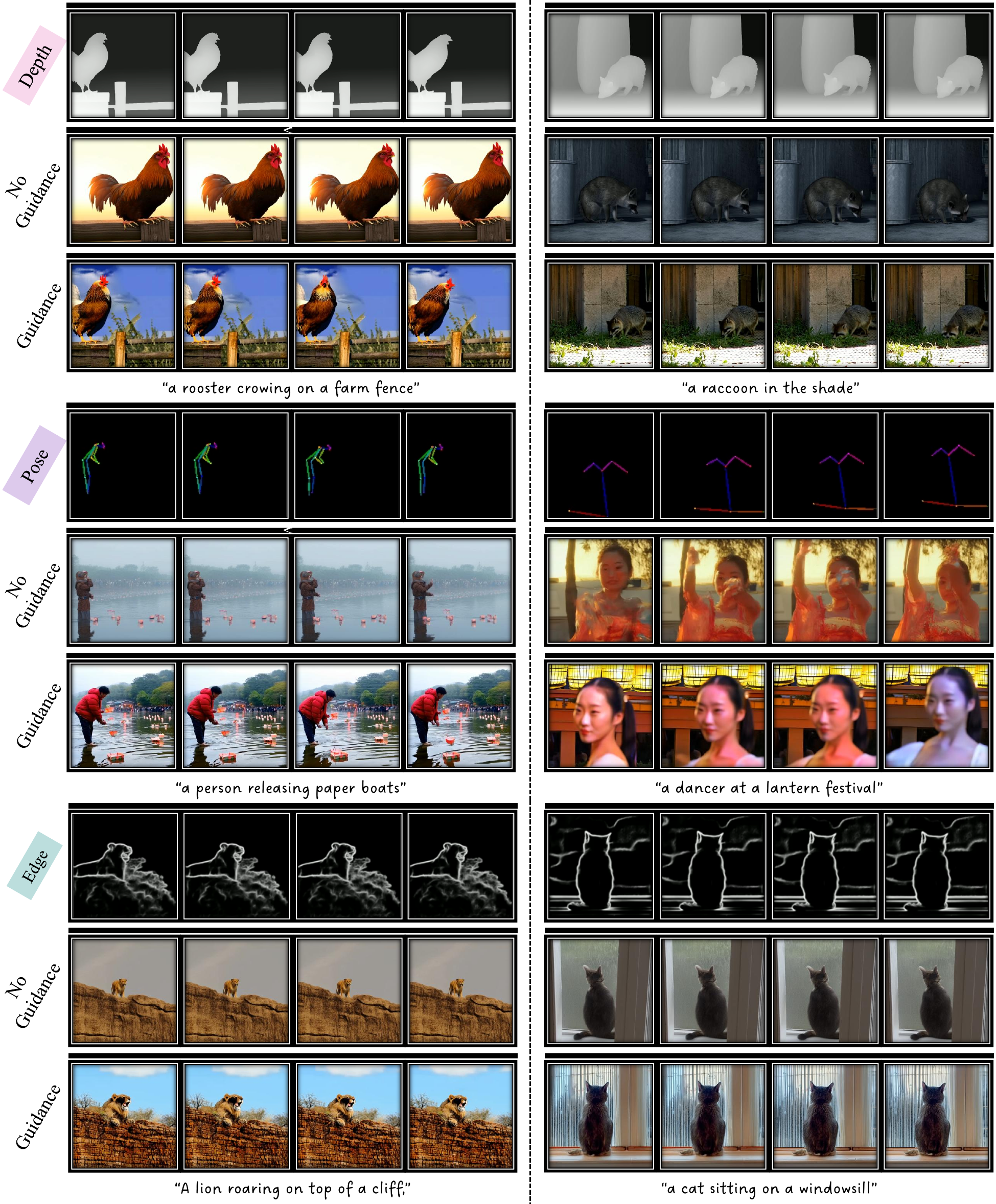}
    \caption{\textbf{Additional Spatial Video Guidance. }Additional examples of spatial video guidance across depth, pose, and edge conditions.
    \name{} successfully guides video generation to follow the target spatial video.}
    \label{fig:suppl_video_all}
    \vspace{-1em}
\end{figure*}

\begin{figure*}[t]
    \centering
    \includegraphics[width=0.7\linewidth]{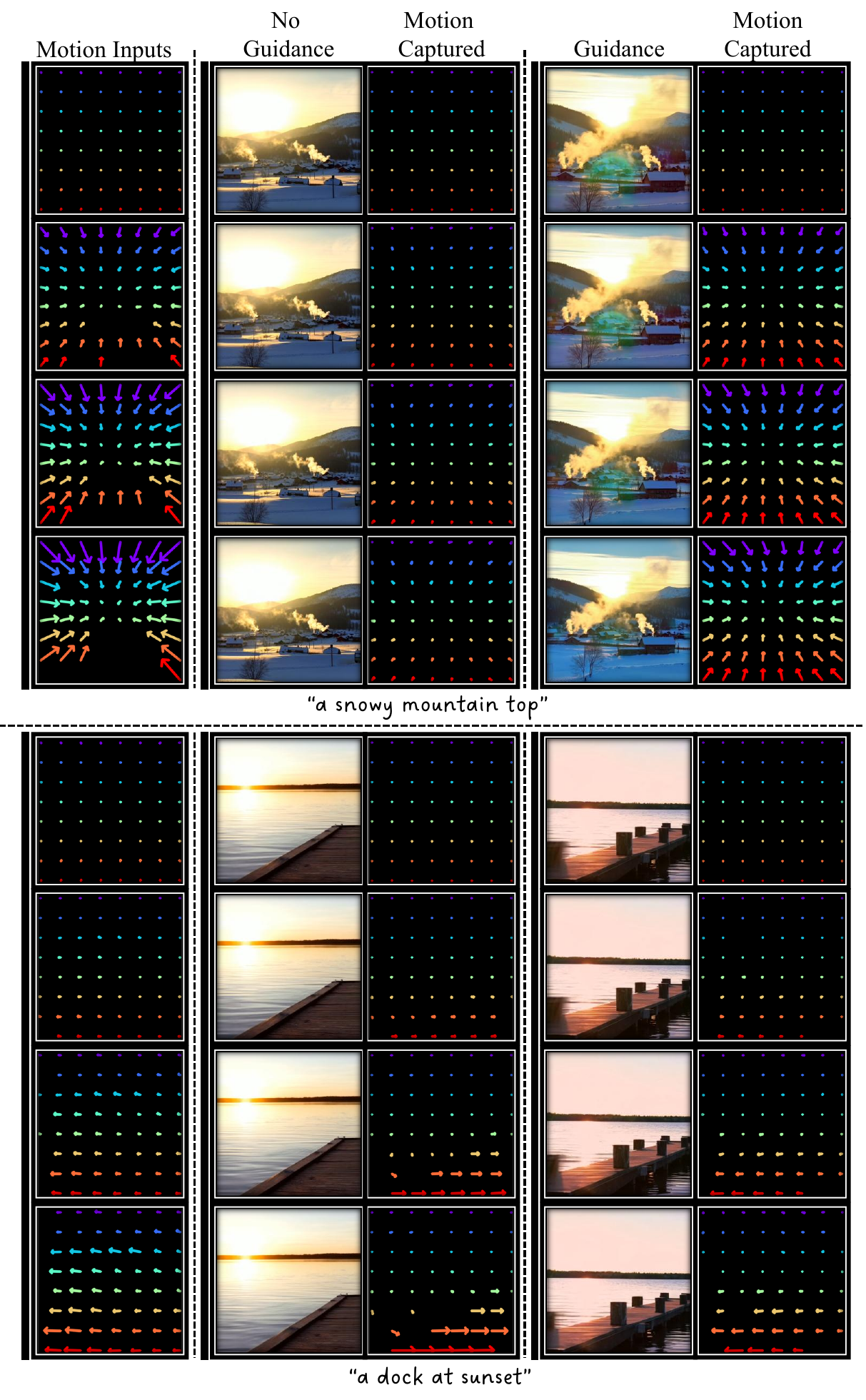}
    \vspace{-1em}
    \caption{\textbf{Additional Motion Video Guidance Results. }Additional examples of motion-guided generation results, showing zoom-in (top) and
pan-right (bottom) camera motions.
    For each example, we present the unguided and guided results alongside their corresponding captured motion. The motion captured in videos guided by \name{} consistently follows the input motion, demonstrating the effectiveness of \name{} in motion-guided video generation.
    }
    \label{fig:suppl_motion}
\end{figure*}

Beyond image generation, \name{} extends naturally to video generation, as shown in Sec.~\ref{sec:video_guidance}. We provide additional spatial video guidance results for depth, pose, and edge in Figure~\ref{fig:suppl_video_all}, and additional motion guidance results in Figure~\ref{fig:suppl_motion}. Compared to unguided generation, \name{} consistently adheres to the input spatial signal while
maintaining temporal consistency and visual quality across all tasks. For visual comparisons, Figure ~\ref{fig:video_sp_compare} shows the performance between RG and our method on spatial tasks.

\subsection{Additional Verification on Diversity and Text Alignment}
\label{sec:suppl_diversity}

By design, \name{} constrains only the modality it is trained to guide, \eg, depth, pose, edge, or motion, and should leave the diversity of the generation and its alignment with the text prompt unchanged. We verify both with two metrics. CLIP-T measures text alignment as the cosine similarity between the CLIP~\cite{radford2021clip} embeddings of a generated image and its prompt. Pairwise LPIPS~\cite{zhang2018lpips} measures diversity as the average LPIPS distance between images generated from the same prompt.

We compare unguided and \name{}-guided generation using the same prompts, seeds, and the guidance control maps as described in Sec.~\ref{sec:eval_setup}. As shown in Table~\ref{tab:diversity}, the two rows are nearly identical under both settings, thus suggesting that \name{} preserves diversity and text alignment while enforcing the target spatial attribute.

\begin{table}[t]
\centering
\setlength{\tabcolsep}{6pt}
\caption{\textbf{Influence of Guidance on Diversity and Text Alignment.}
Pairwise LPIPS and CLIP-T with and without our method, using spatial maps from generated and real images. Scores are nearly identical, indicating that diversity and text alignment are preserved.
}
\renewcommand{\arraystretch}{1}
\scalebox{0.82}{%
    \begin{tabular}{c cc cc}
    \toprule
     & \multicolumn{2}{c}{\textbf{{Generated}}}
     & \multicolumn{2}{c}{\textbf{{Real}}} \\
    \cmidrule(lr){2-3} \cmidrule(lr){4-5}
    {\small \textbf{Method}}
     & {\small \textbf{CLIP-T ($\uparrow$)}} & {\small \textbf{LPIPS ($\uparrow$)}}
     & {\small \textbf{CLIP-T ($\uparrow$)}} & {\small \textbf{LPIPS ($\uparrow$)}} \\
    \midrule
    No Guidance & \textbf{36.14} & \textbf{0.594} & \textbf{32.32} & \textbf{0.647} \\
    Guided (Ours)     & 35.68 & 0.592 & 32.18 & 0.644 \\
    \bottomrule
    \end{tabular}%
}
\label{tab:diversity}
\vspace{-1.5ex}
\end{table}







\section{Guidance Details}
\label{sec:guidance_details}
To optimize the latent effectively, 
multiple optimization steps per denoising step are needed to reach our intended spatial signal. For all tasks, we use 100 denoising steps with AdamW optimizer. The learning rate and denoising steps configurations for various DiT models and spatial tasks are detailed in Table~\ref{tab:guidance_details}. For the first 34 steps, we use 5 refinement steps, for steps 35-49, we use 7 refinement steps, and for steps from 50-60, we use 1 refinement step. Beyond 60 steps, we observe that further optimization does not contribute to guidance quality and is therefore skipped.



\begin{table}[t]
\centering
\caption{\small \textbf{Inference Guidance Details.} Learning rate (LR) schedule relative to the denoising step range.}
\label{tab:guidance_details}
\setlength{\tabcolsep}{5pt}
\small
\begin{tabular}{@{}llcc@{}}
\toprule
\textbf{Base Model} & \textbf{Task} & \textbf{Steps} & \textbf{LR} \\
\midrule
\multirow{2}{*}{SD3-Medium~\cite{stablediffusion3}} & \multirow{2}{*}{Depth, Pose, Edge} & 0--34 & $1.5\!\times\!10^{-3}$ \\
 & & 35--60 & $1.0\!\times\!10^{-3}$ \\
\midrule
\multirow{2}{*}{Flux~\cite{flux2024}} & \multirow{2}{*}{Depth, Pose, Edge} & 0--34 & $3.0\!\times\!10^{-3}$ \\
 & & 35--60 & $1.0\!\times\!10^{-3}$ \\
\midrule
\multirow{7}{*}{CogVideoX~\cite{yang2025cogvideoxtexttovideodiffusionmodels}} & Depth & 0--60 & $2.0\!\times\!10^{-3}$ \\
\cmidrule{2-4}
 & \multirow{2}{*}{Pose} & 0--34 & $1.0\!\times\!10^{-3}$ \\
 & & 35--60 & $5.0\!\times\!10^{-4}$ \\
\cmidrule{2-4}
 & \multirow{2}{*}{Edge} & 0--34 & $1.5\!\times\!10^{-3}$ \\
 & & 35--60 & $1.0\!\times\!10^{-4}$ \\
\cmidrule{2-4}
 & Motion & 0--60 & $1.0\!\times\!10^{-3}$ \\
\bottomrule
\end{tabular}
\end{table}

\section{Evaluation Setup}
\label{sec:eval_setup}

We evaluate all of the models on the MSCOCO \cite{lin2015microsoftcococommonobjects} dataset, specifically the 2017 validation data, and a synthetic dataset. 
The two evaluation settings differ in how the guidance control maps are obtained. In both cases, we run each guidance method conditioned on the prompt $k_i$ and control map $y_i$ to produce a guided image $\hat{x}_i^g$. We then apply the same off-the-shelf spatial predictor to $\hat{x}_i^g$ to obtain the predicted control map $\hat{y}_i$ and compute metrics between $y_i$ and $\hat{y}_i$.

\paragraph{Real control maps (Real). }
For each real image $x_i$, we generate the corresponding prompt $k_i$ using an off-the-shelf Video-Language-Model \cite{Qwen-VL} and the control map $y_i$ using the same task-specific spatial predictor. We treat $k_i$ and $y_i$ as the real image's ground truth prompt and control map. 

\begin{figure}[htb]
    \centering
    \begin{subfigure}[t]{0.48\linewidth}
        \centering
        \includegraphics[width=\linewidth]{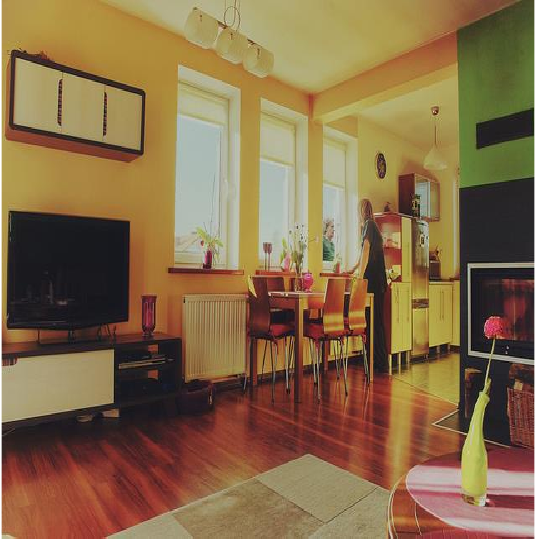}
        \caption{MSCOCO image with generated evaluation prompt: ``The image depicts a cozy living room with a warm, inviting atmosphere, featuring a yellow wall, a wooden dining table with chairs, a flat-screen TV mounted on the wall, and a small plant on a side table.''}
        \label{fig:suppl_caption}
    \end{subfigure}
    \hfill
    \begin{subfigure}[t]{0.48\linewidth}
        \centering
        \includegraphics[width=\linewidth]{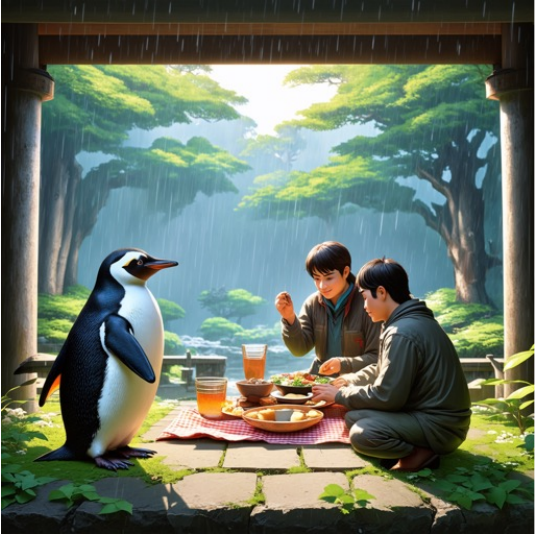}
        \caption{Generated image by SD3-Medium with prompt: ``A adelie penguin sharing a picnic with explorers inside an ancient forest temple in Kyoto during a summer storm, Realistic Style.''}
        \label{fig:suppl_caption_2}
    \end{subfigure}
    \caption{\textbf{Examples of Evaluation Images and Prompts.} (a) A real MSCOCO image with its VLM-generated caption. (b) A generated image from SD3-Medium with its text prompt.}
    \label{fig:eval_examples}
\end{figure}

\paragraph{Generated control maps (Gen). } 
We generate 250 diverse prompts $k_i$ using ChatGPT, covering a range of scenes, objects, animals and compositions in a realistic style. Each prompt $k_i$ is fed into SD3 \cite{stablediffusion3} to produce an unguided image $\hat{x}_i$, from which we extract the control map $y_i$ using the corresponding off-the-shelf spatial predictor (e.g, DepthAnythingV2 for depth, OpenPose for pose, HED for edge). This control map $y_i$ serves as the guidance target. Figure~\ref{fig:eval_examples} shows both evaluation pipelines with example images and their corresponding prompts.

\paragraph{Video motion guidance evaluation.}
For motion guidance, we first generate 6 reference motions from CogVideoX ~\cite{yang2025cogvideoxtexttovideodiffusionmodels} (zoom in, zoom out, right, left, up, and down). From here, we generate 20 prompts using ChatGPT covering a range of scenes, objects, and compositions in a realistic style. For each prompt, we randomly select a motion and feed a prompt with our selected motion into our setup.

\paragraph{Video spatial guidance evaluation.}
For spatial guidance, we generate over 100 prompts from ChatGPT covering a range of scenes, objects, animals and compositions in a realistic style. Each prompt $k_i$ is fed into CogVideoX \cite{stablediffusion3} to produce an unguided video $\hat{x}_i$, from which we extract the control maps $y_i$ using the corresponding off-the-shelf spatial predictor (e.g, DepthAnythingV2 for depth, OpenPose for pose, HED for edge). These control maps $y_i$ serve as the guidance target.

\section{Prompt Details}

In our experiments, we use various prompts covering different scene complexity styles, and semantic categories. Specifically, the results shown in Figure~\ref{fig:spatial_guidance} use prompts of the same format
but describing different actions and styles to showcase the stability and
generalization of our method. The full prompts used in this figure are (from top to bottom, left to right, column by column): 

\begin{itemize}
    \item A white paper-cut sculpture of a standing philosopher in a uniform studio.
    \item A humanoid made of polished liquid chrome, standing in a uniform studio.
    \item A white marble sculpture of a standing man in a uniform studio.
    \item A woman in a dojo uniform is sweeping the floor in a dojo at night time.
    \item A woman in a sportswear is playing and passing a volleyball in a volleyball court.
    \item A female student wearing a white t-shirt and jeans is sweeping the floor in an empty studio.
    \item A doctor in a white lab coat, sitting in a minimalist chair, holds a white ceramic mug in a light-blue walled studio.
    \item A woman in pajamas is reading a newspaper and sitting in an armchair in a bedroom.
    \item An actor in a suit is holding a coffee cup and sitting in a wooden chair in a studio.
\end{itemize}

These prompts are deliberately selected to span a range of object materials, human actions, lighting conditions, and environments to ensure a comprehensive evaluation of \name{}'s generalization capability.

\section{Asset Licenses and Terms of Use}
The pre-trained models used in this work are released under the following licenses: CogVideoX is licensed under Apache 2.0, while Stable Diffusion 3 and FLUX.1 [dev] operate under custom non-commercial research licenses. The official repositories and model cards, which contain the full license agreements, are provided in our references.




\end{document}